\documentclass{article}

\PassOptionsToPackage{numbers, compress}{natbib}

 \usepackage[main, final]{neurips_2026}

\usepackage[utf8]{inputenc} 
\usepackage[T1]{fontenc}    
\usepackage[hidelinks,pagebackref, colorlinks]{hyperref}
\usepackage{url}            
\usepackage{booktabs}       
\usepackage{amsfonts}       
\usepackage{nicefrac}       
\usepackage{microtype}      
\usepackage{xcolor}         
\usepackage{caption}
\usepackage{adjustbox}
\definecolor{customcolor}{HTML}{DA335C}
\definecolor{citecustomcolor}{HTML}{2B6691}
\hypersetup{
    colorlinks=true,
    linkcolor=customcolor,
    filecolor=customcolor,
    urlcolor=customcolor,
    citecolor=citecustomcolor
}

\usepackage{amsmath}
\usepackage{amssymb}
\usepackage{mathtools}
\usepackage{amsthm}
\usepackage{siunitx}
\usepackage{multirow}
\usepackage[most]{tcolorbox}
\usepackage{tabularx}
\usepackage{pdflscape}
\usepackage{array}
\usepackage{makecell}

\usepackage[capitalize,noabbrev]{cleveref}
\usepackage[table, svgnames]{xcolor} 
\usepackage{xspace}
\usepackage{enumitem}
\usepackage{amsmath}
\usepackage{amssymb}
\usepackage{float}
\newcommand{\blue}[1]{\textcolor{blue}{#1}}
\newcommand{\red}[1]{\textcolor{red}{#1}}
\usepackage[capitalize,noabbrev]{cleveref}
\usepackage{wrapfig}

\usepackage{amsmath}

\theoremstyle{plain}

\theoremstyle{definition}

\theoremstyle{remark}

\newcommand{\gctsg}{\text{GC-TSG}\xspace}

\newcommand{\name}{\texttt{TimeTok}\xspace}
\newcommand{\define}[1]{\vspace{0mm}\noindent{{\textbf{#1.}}}}
\newcommand{\ie}{\emph{i.e.,}\xspace}
\newcommand{\eg}{\emph{e.g.,}\xspace}
\newcommand{\std}[1]{\scriptsize{$\pm$#1}}
\newcommand{\stdtiny}[1]{\tiny{$\pm$#1}}
\newcommand{\defeq}{\overset{\text{\tiny def}}{=}}
\newcommand{\ldefeq}{\mathrel{\raisebox{-0.3ex}{$\defeq$}}}

\newcommand{\disteq}{\overset{\text{\tiny dist.}}{=}}
\newcommand{\ldisteq}{\mathrel{\raisebox{-0.3ex}{$\disteq$}}}

\definecolor{mygreybox}{HTML}{F1F3F8}

\definecolor{greydark}{RGB}{90, 90, 90} 
\definecolor{greylight}{HTML}{F1F3F8}

\newtcolorbox{greycustomblock}{
  colframe=greydark,        
  colback=greylight,        
  boxrule=1pt,              
  left=2.5pt,               
  right=3pt,                
  top=5pt,                  
  bottom=3pt,               
  arc=0pt,                  
  breakable,                
  before skip=0.2\baselineskip, 
  after skip=0.2\baselineskip,  
  left skip=0pt,            
  right skip=0pt,           
  enhanced jigsaw,          
  frame hidden,             
  overlay={                 
    \draw[greydark, line width=2pt]
      ([yshift=-1pt]frame.north west) -- ([yshift=1pt]frame.south west); 
  },
  fontupper=\selectfont, 
}

\usepackage{xcolor}
\usepackage[most]{tcolorbox}
\usepackage{tikz}
\usepackage{fontawesome}

\definecolor{insightteal}{RGB}{34, 139, 139}   
\definecolor{insightback}{RGB}{240, 248, 248}   

\newcommand{\lightbulbicon}{%
  \begin{tikzpicture}[baseline=-0.5ex]
    \draw[fill=white, draw=insightteal, thick] (0,0) circle (1.5ex);
    \node[scale=0.8, color=insightteal] at (0,0) {\faLightbulbO~};
  \end{tikzpicture}%
}

\newtcolorbox{customblockquote}{
  colframe=insightteal,
  colback=insightback,
  boxrule=0pt,
  left=5pt,
  right=4pt,
  top=5pt,
  bottom=3pt,
  arc=0pt,
  breakable,
  before skip=1.2\baselineskip,
  after skip=0.7\baselineskip,
  left skip=0pt,
  right skip=0pt,
  enhanced jigsaw,
  frame hidden,
  overlay={
    \draw[insightteal, line width=2pt] 
      ([yshift=1pt]frame.north west) -- (frame.south west);
    \node[inner sep=0pt] at ([xshift=0pt, yshift=-1.3pt]frame.north west) {\lightbulbicon};
  },
  fontupper=\fontfamily{lmr}\selectfont,
  boxsep=1pt,
}

\usepackage[textsize=tiny]{todonotes}

\title{\name: Granularity-Controllable Time-Series Generation via Hierarchical Tokenization}

\author{%
  Seokhyun Lee\textsuperscript{1}\thanks{Equal contribution.}\quad
  Jaeho Kim\textsuperscript{1}\footnotemark[1]\quad
  Changjun Oh\textsuperscript{1}\\
  \textbf{Mihaela van der Schaar}\textsuperscript{2}\quad
  \textbf{Changhee Lee}\textsuperscript{1}%
  \thanks{Corresponding author: \texttt{changheelee@korea.ac.kr}}\\
  \textsuperscript{1}Department of Artificial Intelligence,
  Korea University, Seoul, South Korea\\
  \textsuperscript{2}Department of Applied Mathematics and Theoretical Physics,\\
  University of Cambridge, Cambridge, United Kingdom
}

\begin{document}

\vspace{-15pt}
\maketitle

\vspace{-20pt}
\begin{abstract}
\vspace{-10pt}
Time-series data are inherently multiscale, spanning diverse temporal granularities from coarse trends to fine-scale dynamics. However, existing time-series generative models provide limited control over the temporal granularity of both inputs and outputs, restricting their ability to condition on user-provided coarse sketches and generate samples at a desired target granularity. To address this, we introduce \name, a unified framework for \emph{Granularity-Controllable Time-Series Generation}~(\gctsg), which generates time series at \emph{any} target granularity from \emph{any} coarser input~(\eg rough sketches) or without conditioning. At the core of \name is a hierarchical tokenization strategy that maps time series into an ordered sequence of tokens, from coarse to fine temporal granularity. Our autoregressive generation process operates across these granularity levels, producing token blocks that are decoded back into continuous time series. This design naturally enables \gctsg within a single framework, where controlling the number of token blocks provides explicit control over output detail. Experiments show that \name excels at \gctsg tasks while achieving state-of-the-art performance in standard generation. Furthermore, we showcase \name's potential as a foundational tokenizer by training on multiple datasets with heterogeneous temporal granularities, verifying strong transferability that consistently outperforms models trained on individual datasets. To our knowledge, this is the first unified framework that covers the full generative spectrum for time series, offering a valuable foundation for models that benefit from diverse temporal granularities. Codes are available \href{https://github.com/nshuhsn/TimeTok_NeurIPS.git}{\underline{here}}.
\end{abstract}
\vspace{-10pt}
\section{Introduction}
\label{sec:intro}

Time-series data are inherently multiscale: they can be observed, represented, and analyzed at different levels of temporal granularities. Coarse-grained representations capture slowly varying trends and long-range dependencies, whereas fine-grained representations preserve rapid fluctuations and detailed dynamics. In practice, however, acquiring fine-grained time-series data is often costly or impractical due to hardware costs, storage limitations, and privacy concerns. As a result, there is a persistent granularity mismatch between the coarse-grained data typically available and the fine-grained information required for reliable downstream modeling and decision-making. 

This mismatch exposes a limitation of the standard formulation of time-series generation. Existing models typically learn to sample time series at the single temporal granularity present in the training data, but they do not provide a native interface for asking: given this coarse observation, what fine-grained dynamics are plausible? Or, given no observation, can we generate directly at a user-specified level of temporal detail? Thus, the missing capability is not only better sample quality, but also \emph{controllability over the temporal granularity of both the conditioning input and the generated output.} 

This challenge is particularly consequential in high-stakes domains such as healthcare and finance, where decision-making and data analysis routinely span multiple temporal granularities. In healthcare, hospital-grade electrocardiogram~(ECG) signals are recorded at high sampling rates (up to 1000 Hz) but consumer-grade smartwatches collect ECG data at much lower sampling rates~(often below 60 Hz)~\citep{isakadze2020useful}. In finance, market data is recorded at a millisecond level but analyzed at task-dependent granularities, ranging from high-frequency trading~\citep{moreno2024deepvol} to monthly portfolio rebalancing~\citep{hwang2025deep}, yet existing time-series generation methods provide limited support for bridging across these scales. 


We therefore study granularity-controllable time-series generation (\gctsg): the problem of learning \emph{one generative model that can synthesize time series at any target granularity, either unconditionally or conditioned on an input observed at any coarser granularity.} Standard time-series generation is then only one special case, corresponding to generation from scratch at the finest granularity. Conditional refinement, such as transforming a low-fidelity ECG or a user-provided stock-market sketch into a plausible fine-grained trajectory, is another case of the same problem. This framing turns temporal granularity from a nuisance of preprocessing into an explicit generative control variable. The central technical question is therefore \emph{how can we build a time-series representation in which temporal granularity is ordered, controllable, and generative.}


To answer this, we propose \name, a unified framework for granularity-controllable time-series generation. At the core of \name is a hierarchical tokenizer that represents a time series as an ordered sequence of discrete tokens, where early tokens encode coarse temporal structure and later tokens add progressively finer dynamics. This gives token length a semantic meaning: choosing how many tokens to decode controls the temporal granularity of the output. Building on this representation, \name uses block-wise autoregressive generation across granularity levels, generating the next block of finer-grained tokens conditioned on the coarser prefix. The resulting model can generate from scratch at any granularity, refine coarse observations into finer trajectories, and use user-provided sketches as conditioning inputs, all within the same architecture.

\begin{customblockquote}
\define{Contribution} (i) We formulate \gctsg as a unified generative problem over pairs of temporal granularities, making standard generation and conditional temporal refinement instances of the same task. (ii) We introduce \name, a hierarchical tokenization and autoregressive generation framework in which temporal granularity is controlled directly through the number of generated token blocks. (iii) We show empirically that this representation is not only controllable but useful: \name outperforms strong time-series generation baselines on \gctsg, achieves state-of-the-art or competitive standard generation performance, improves downstream prediction under granularity mismatch, and scales to a foundation-tokenizer trained on heterogeneous multi-domain time-series data. 
\end{customblockquote}



\begin{figure*}[t]
\begin{center}
\center{\includegraphics[width=0.97\textwidth]{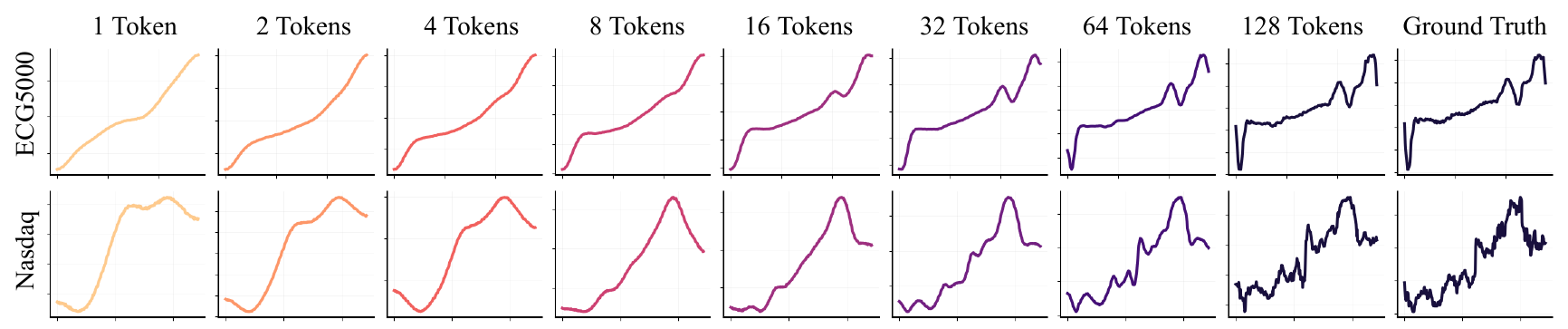}}
\caption{\small\define{Visualization at Different Granularity} \name encodes time series into a hierarchically ordered sequence of discrete tokens that progressively capture coarse-to-fine temporal granularities. We visualize samples from the medical~(\ie ECG) and finance domain~(\ie Nasdaq stock data); additional results in~\cref{appendix:full_coarse_2_fine}. }
\vspace{-10pt}
\label{fig:coarse_to_fine_recon}
\end{center}
\end{figure*}

\vspace{-5pt}
\section{Related Works}\label{sec:rel}
\vspace{-5pt}

\begin{table*}[t]
\centering
\small
\caption{
Contributions of \name relative to prior works. More detailed comparisons are provided in~\cref{tab:related_work_comparison_rotated}.}
\label{tab:related_work_positioning}
\resizebox{\textwidth}{!}{
\begin{tabular}{p{0.12\textwidth}p{0.16\textwidth}p{0.25\textwidth}p{0.51\textwidth}}
\toprule
\textbf{Line of work} &
\textbf{Methods} &
\textbf{Main capability} &
\textbf{Limitation in \gctsg / Relation to \name}\\
\midrule

Standard time-series generation&

TimeGAN~\citep{yoon_timegan_2019NeurIPS}, TSGDiff~\citep{shen_tsgdiff_AAAI2026}, TimeVQVAE~\citep{Daesoo_TimeVQVAE_2023AISTATS}, Diffusion-TS~\citep{yuandiffusion}, ImagenTime~\citep{naiman_ImagenTime_NeurIPS2024}&
Generates quality samples at the \emph{trained} granularity.&
These models do not natively expose temporal granularity as a controllable dimension or model transitions between granularities. \name includes Standard time-series generation as a special case,  but also supports conditional refinement and generation at any target granularity\\
\midrule
Time-series tokenization&
TOTEM~\citep{talukdertotem}&
Learns general-purpose token representations for time-series tasks.&
Token representations from these models are unordered. \name imposes an ordered temporal hierarchy over tokens, enabling controllable granularity. \\
\midrule
Multi-granular tokens&

WaveToken~\citep{masserano_wavetok_2025ICML}, TransPL~\citep{kimtranspl}&
Captures frequency-aware and trend--residual temporal structure.&
Granularity is usually fixed by the architecture rather than specified by the user at inference time. \name turns granularity into a generative control variable.\\
\midrule
Hierarchical image generation&

VAR~\citep{tian_VAR_2024NeurIPS}, FlowAR~\citep{renflowar}, FlexTok~\citep{bachmann_flextok_2025ICML}&
Uses ordered or variable-length token generation to improve image synthesis quality and efficiency.&
Intermediate scales are mainly latent scaffolds for the final-resolution image, and variable prefixes control the complexity of generated image rather than modeling refinement dynamics or conditional generation across granularity pairs. \name instead treats granularity as user-facing input-output control for conditional refinement.\\

\midrule
\rowcolor[HTML]{F0F8F8} 
\textbf{\name}&
\textbf{This work} &
\textbf{Hierarchical temporal tokenization with block-wise autoregressive coarse-to-fine generation.} &
\textbf{\name is the first unified framework that enables: (i) time series generation from scratch, (ii) generation at intermediate granularities, and (iii) conditional refinement from coarse inputs.}\\

\bottomrule
\end{tabular}}\end{table*}

\name connects three lines of work that have been mostly discussed separately: time-series generation, time-series tokenization, and coarse-to-fine token generation. Its novelty is not any one component in isolation, but the alignment of these three components into a single, unified generative interface for \gctsg. The hierarchical tokenizer gives token prefixes a temporal-granularity semantics; the block-wise autoregressive model learns a refinement trajectory across those prefixes; and the inference procedure allows arbitrary coarse inputs, including low-fidelity observations and user sketches, to be refined at a user-defined granularity. This enables \name to support standard generation, unconditional generation towards any intermediate granularities, and conditional coarse-to-fine refinement, all within a single model. We summarize the contributions of \name relative to prior works in~\cref{tab:related_work_positioning} and provide a more detailed discussion of the related works in~\cref{app:related_work}.

\section{
Modeling Time-Series Generation Across Temporal Granularities}\label{sec:problem}

In this work, we posit that a time-series generative model should treat \textit{granularity} as a controllable dimension, enabling traversal across granularity levels during generation.

\subsection{Generation with Multiple Temporal Granularity}
Let $\mathbf{x} = [x_1,\dots,x_T] \in \mathbb{R}^{T}$ be a univariate time series of length $T$, drawn from an unknown data distribution $p_{\text{data}}$, \ie $\mathbf{x} \sim p_{\text{data}}(\mathbf{x})$. 
We formally define granularity level $\ell \in \{1,\dots,L\}$ as a specification of how much temporal detail is preserved in a given time series. 
For each time series~$\mathbf{x}$, we denote $\mathbf{x}^{(\ell)}$ as a time series with temporal granularity level $\ell$ where each $\mathbf{x}^{(\ell)}$ shares the same high-level semantics as $\mathbf{x}$ but differs in its temporal details; a smaller value corresponds to coarser time series, while a larger value preserves more fine-grained temporal information. By convention, we set $\mathbf{x}^{(L)} \ldefeq \mathbf{x}$, corresponding to the finest granularity and indicating the original time series.


\textbf{Standard Time-series Generation.}~
The \textit{Standard time-series generation} focuses on learning a density~$p_\theta$ whose samples are indistinguishable from those drawn from the marginal data distribution~$p_{\text{data}}$, achieving $p_{\theta}(\mathbf{x}) \ldisteq p_{\text{data}}(\mathbf{x})$, where $\ldisteq$ denotes distributional equality. 
While fundamental, it offers no control over the temporal detail of outputs or conditioning on high-level temporal structures. 

\textbf{Granularity-Controllable Time-series Generation.}~
We define \gctsg as a generalized time-series generation task with explicit control over temporal granularity for both conditioning and synthesis as follows:

\begin{greycustomblock}
\paragraph{Definition (\gctsg).}
Let ${\mathbf{x}}^{(i)}$ and ${\mathbf{x}}^{(j)}$ denote time-series representations at temporal granularity levels $i, j \in \{0,1,\dots, L\}$ with $j > i$. 
We define \textit{granularity-controllable time-series generation (\gctsg)} as the task of learning a generative model $p_{\theta}$ whose samples are indistinguishable from those drawn from the conditional data distribution across all valid pairs of temporal granularities, achieving $p_\theta({\mathbf{x}}^{(j)} \mid {\mathbf{x}}^{(i)}) \ldisteq p_{\text{data}}({\mathbf{x}}^{(j)} \mid {\mathbf{x}}^{(i)})$. 
Here, we slightly abuse notation by defining $\mathbf{x}^{(0)} = \varnothing$, which corresponds to the unconditioned setting.
\end{greycustomblock}

\begin{figure*}[t]
\begin{center}
\center{\includegraphics[width=0.97\textwidth]{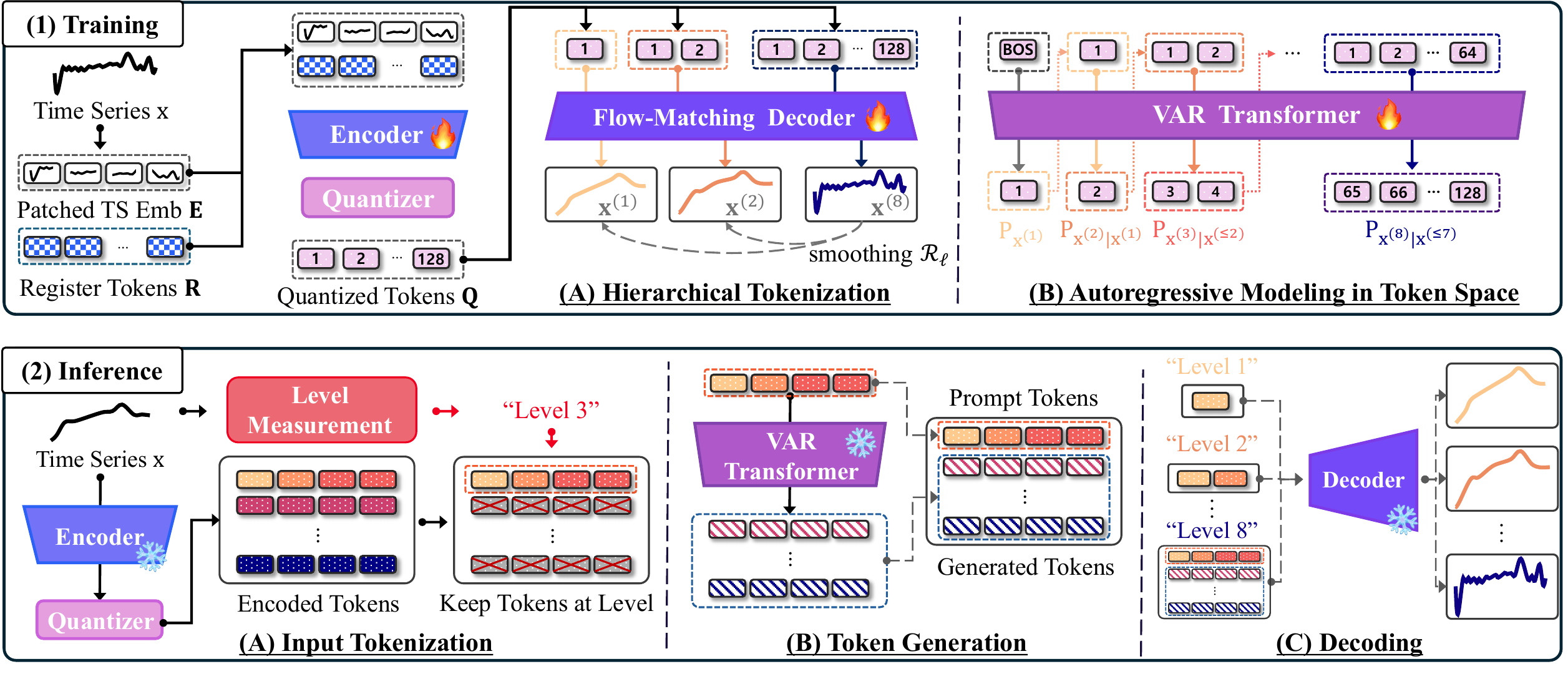}}  \vspace{-1.2mm}
    \caption{\small\define{An Overview of \name} \textbf{(1) Training:}  The hierarchical tokenizer is trained to encode input time series into a sequence of discrete tokens structured in a coarse-to-fine temporal hierarchy. These tokens are then used to train the VAR Transformer that autoregressively predicts tokens at the next granularity level. 
    \textbf{(2)~Inference:} A conditioning time series of unknown granularity is encoded into a full token sequence, from which tokens at the inferred granularity are selected as a prefix for the VAR Transformer to generate the remaining tokens. The completed token sequence is then decoded into a time series at the target granularity.} \vspace{-3.5mm}
    \label{fig:overview}
\vspace{-5pt}
\end{center}
\vspace{-1pt}
\end{figure*}

This formulation unifies unconditional generation (\ie $i=0$) and conditional refinement (\ie $i \geq 1$) under a single framework. Standard generation corresponds to the special case $i=0$ and $j=L$.


\subsection{Autoregressive Formulation of \gctsg}
In practice, time-series datasets typically provide only the finest-grained observations $\mathbf{x}^{(L)}$. To construct the multi-granular hierarchy required for granularity-controllable time-series generation in a principled manner, we introduce a family of \textit{information-reducing operators} $\{ \mathcal{R}_\ell \}_{\ell \in [L]}$. Each operator $\mathcal{R}_\ell$ maps the raw time series to its representation at temporal granularity level $\ell$, \ie $\mathbf{x}^{(\ell)} = \mathcal{R}_\ell(\mathbf{x})$. Here, smaller values of $\ell$ induce stronger information reduction, resulting in coarser abstractions of the time series. For instance, $\mathcal{R}_{\ell}$ can be implemented as a Gaussian smoothing kernel~\citep{zhao_trendestimation_2003Elsevier} or a moving-average operator, where $\ell$ controls the smoothing variance or the window size, respectively, to determine the temporal granularity\footnote{In practice, the degree of temporal granularity can be quantified using established measures in time series, such as the detrended fluctuation analysis (DFA) exponent~\citep{peng_DFA_1995AIP}.}.

A core objective of \gctsg is to model conditional distributions over time-series representations for any valid pair of temporal granularities, using only the finest-grained time series (\ie observations). To achieve this, we decompose the marginal log-likelihood of the observed time series by applying the chain rule along the granularity axis:
\vspace{-3pt}
\begin{align} \label{eq:Chain_Rule}
\log p_\theta (\mathbf{x} ) \! = \log p_\theta\big(\mathbf{x}^{(\le L)}\big) 
    = \!  {\textstyle\sum_{\ell=1}^{L}} \log p_\theta\big({\mathbf{x}}^{(\ell)} \!\!\mid\! {\mathbf{x}}^{(<\ell)}\big)\!
\end{align} 
\vspace{-3pt}
where $\mathbf{x}^{(<\ell)} = \big( \mathbf{x}^{(1)}, \dots, \mathbf{x}^{(\ell-1)} \big)$. The first equality holds because each coarse-granular representation $\mathbf{x}^{(\ell)}$ is deterministically obtained from $\mathbf{x}$ via $\mathcal{R}_{\ell}$. 
This hierarchical autoregressive decomposition induces a joint distribution over representations at all temporal granularities and, in doing so, defines a valid \textit{refinement process} that incrementally introduces finer temporal details. 

Under this formulation, generating $\mathbf{x}^{(j)}$ conditioned on $\mathbf{x}^{(i)}$ is realized by sampling a coarse-to-fine sub-trajectory. Conditioning on a prefix consistent with $\mathbf{x}^{(i)}$, ancestral sampling proceeds through successive conditional distributions $p_\theta\big({\mathbf{x}}^{(\ell)} \!\!\mid\! {\mathbf{x}}^{(<\ell)}\big)$ for $\ell = i+1, \dots, j$, progressively refining the time-series representations until reaching granularity level $j$. 
This procedure ensures that the generated $\mathbf{x}^{(j)}$ remains both statistically consistent with the conditioning input $\mathbf{x}^{(i)}$ and coherent with the hierarchical structure.

Unfortunately, learning and sampling from this hierarchical formulation in Eq.~\eqref{eq:Chain_Rule} present two key challenges: 
(i) Implementing each refinement step -- \ie $p_{\theta}(\mathbf{x}^{(\ell)}|\mathbf{x}^{(<\ell)})$ -- directly in the raw time-series space is inherently difficult, as it requires conditioning on the entire trajectory of preceding coarser time series, which makes training computationally expensive and often unstable, especially for long time series. 
(ii) Conditional generation given $\mathbf{x}^{(i)}$ implicitly assumes access to the full coarse-level prefix $\mathbf{x}^{(\le i)}$, which is typically unavailable at inference time and otherwise needs to be reconstructed. These challenges motivate hierarchical modeling in a structured discrete token space.


\section{Method: \texttt{\name} } \label{sec:tokenization}

To address the challenge of modeling and generating time series with hierarchical dependencies across temporal granularities, we introduce \name, a novel framework for GC-TSG comprising three key components; an overview of \name is illustrated in Figure \ref{fig:overview}:  \vspace{-0.1em}
\begin{itemize}[leftmargin=1.2em, topsep=0pt, partopsep=0pt, itemsep=0pt]
    \item \textbf{Hierarchical Tokenization.} Continuous time series are mapped into a sequence of discrete tokens that encode temporal structure in a coarse-to-fine hierarchy. \vspace{-0.2em}
    \item \textbf{Autoregressive Modeling in Token Space.} Generative modeling is then performed in the learned discrete token space using a hierarchical autoregressive model. \vspace{-0.2em}
    \item \textbf{Granularity-Controllable Inference.} At inference, coarse time-series structure is encoded into a token prefix, and additional tokens are autoregressively generated up to a target granularity. The resulting token sequence is decoded to produce a time-series at the desired temporal granularity. \vspace{-0.2em}
\end{itemize}

\subsection{Hierarchical Tokenization for Granularity Control}

To encode a coarse-to-fine temporal hierarchy directly in the discrete latent space, we introduce a structured tokenization in which the token ordering is semantically aligned with temporal granularity. Specifically, the leading tokens are constrained to represent high-level information of the time series, while successive tokens incrementally encode fine-scale temporal dynamics. In contrast to conventional tokenization schemes that require generating a fixed-length token sequence, our approach allows variable-length token sequences depending on the temporal granularity.

\define{Discrete Representation via Register Tokens}
Following the discrete representation learning paradigm~\citep{van_VQvae_2017NeurIPS, mentzer_FSQ_2023arXiv}, we assume an encoder--decoder architecture $(\mathtt{Enc}, \mathtt{Dec})$ and a quantizer $\mathcal{Q}$. Given an input time series $\mathbf{x}$, we first partition it into $W$ non-overlapping patches of equal length. Each patch is linearly projected into a latent space, producing a sequence of patch embeddings $\mathbf{E} = [\mathbf{e}_1, \dots, \mathbf{e}_W] \in \mathbb{R}^{W \times h}$, where $h$ denotes the hidden dimension. To obtain a compact and structured representation, we append $M$ \textit{learnable} register tokens $\mathbf{R} = [\mathbf{r}_1, \dots, \mathbf{r}_M] \in \mathbb{R}^{M \times h}$~\citep{darcet_register_2024ICLR} to the patch embeddings, forming the augmented sequence $\mathbf{E} \oplus \mathbf{R} \in \mathbb{R}^{(W+M) \times h}$, where~$\oplus$ denotes concatenation. This sequence is then processed by a Transformer-based encoder ($\mathtt{Enc}$). 
We denote the encoder outputs corresponding to the register tokens as $\widetilde{\mathbf{R}} = [\tilde{\mathbf{r}}_1, \dots, \tilde{\mathbf{r}}_M]$. Each encoded register token is subsequently discretized by finite scalar quantization (FSQ)~\citep{mentzer_FSQ_2023arXiv}, resulting in a sequence of discrete latent codes. 
In our implementation, we configure the FSQ codebook with levels \texttt{[4,4,4,4,4,4]}, which maps each register token to a 6-dimensional discrete code, inducing an effective vocabulary size of $4^6$. The resulting quantized register tokens $\mathbf{Q}= [\mathbf{q}_1, \dots, \mathbf{q}_M] = [\mathcal{Q}(\tilde{\mathbf{r}}_{1}), \dots, \mathcal{Q}(\tilde{\mathbf{r}}_{M})] \in \mathbb{R}^{M\times 6}$ are used as the conditioning input to the decoder.

\define{Token Conditioned Flow Matching Decoder} To enforce the desired hierarchical structure, we design the decoder~($\mathtt{Dec}$) using a \emph{Conditional Flow Matching} framework~\citep{lipman_flowmatching_2023ICLR}. Flow Matching generates samples by continuously transforming a source distribution (\eg Gaussian noise) into the data distribution via a time-dependent velocity field. Our key idea is to endow the quantized register tokens with explicit hierarchical semantics by training the decoder to reconstruct time-series representations at different temporal granularities, depending on how many tokens are provided.

For each target granularity level $\ell$, we define $n_\ell$ as the number of tokens required to represent the time series at that level, with $n_L = M$. In practice, we set $n_\ell = 2^{\ell-1}$ and fix the maximum granularity level to $L = 8$. Empirically, we observe that uniform and exponential token allocations produce nearly identical losses across levels, indicating that fewer tokens suffice for coarser levels~(\cref{app:Ablation}). The decoder is conditioned only on the $n_{\ell}$ leading tokens $\mathbf{Q}_{1:n_{\ell}} = [\mathbf{q}_{1}, \dots, \mathbf{q}_{n_{\ell}}]$ and tasked with generating the corresponding time-series representation $\mathbf{x}^{(\ell)}$. 
Here, the target $\mathbf{x}^{(\ell)}$ is obtained from the original time-series $\mathbf{x}$ via the information-reducing operator $\mathcal{R}_{\ell}$, \ie $\mathbf{x}^{(\ell)} = \mathcal{R}_{\ell}(\mathbf{x})$.\footnote{In our implementation, we instantiate $\mathcal{R}_{\ell}$ as a Gaussian smoothing operator that progressively removes fine-grained temporal details. See more details provided in Appendix~\ref{Appendix-subsec:Gaussian-kernel-smoothing}.} In contrast to conventional decoders, which condition the velocity field on the full token sequence $\mathbf{Q}$, we train the velocity predictor $\mathbf{v}_\theta$~(parameterized by a neural network) under varying conditioning budgets, corresponding to different numbers of available tokens. Each conditioning budget induces a distinct generative trajectory: conditioning on fewer tokens encourages the model to capture coarse, high-level temporal structure, while additional tokens progressively refine generation with finer details.

Formally, for each granularity level $\ell$, we define the linear interpolation between a clean sample $\mathbf{x}^{(\ell)}$ and noise $\boldsymbol{\epsilon}\!\sim\!\mathcal{N}(\mathbf{0}, I)$ at time $\tau \in [0,1]$ as $\mathbf{x}^{(\ell)}_\tau = (1 - \tau)\,\boldsymbol{\epsilon} + \tau\,\mathbf{x}^{(\ell)}, $
with the corresponding target velocity $\mathbf{v}_{\tau}^{(\ell)} = \mathbf{x}^{(\ell)} - \boldsymbol{\epsilon}$. This velocity specifies the instantaneous direction that transports noise toward $\mathbf{x}^{(\ell)}$, and is approximated by the parameterized velocity $\mathbf{v}_{\theta}$ conditioned on $\mathbf{Q}_{1:n_\ell}$. 
We train $\mathbf{v}_{\theta}$ using the following coarse-to-fine flow-matching objective:
\begin{equation}
\label{eq:ctf}
\mathcal{L}_{\mathrm{CTF}}(\theta) =
\mathbb{E}_{\mathbf{x},\, \boldsymbol{\epsilon},\, \tau,\, \ell}
\Big[ \, \big\| \mathbf{v}_\theta(\mathbf{x}_\tau^{(\ell)} \! \mid \! \mathbf{Q}_{1:n_\ell}, \tau) - \mathbf{v}_\tau^{(\ell)} \big\|_2^2 \, \Big],\!
\end{equation}
where the granularity level $\ell$ is sampled uniformly during training. This objective explicitly enforces that conditioning on only the first $n_\ell$ tokens suffices to reconstruct the time-series representation with granularity level $\ell$, thus inducing a coarse-to-fine semantic ordering over the token sequence.

\subsection{GC-TSG through Autoregressive Transformer}
\label{sec:generation}

Once the encoder and decoder $(\mathtt{Enc}, \mathtt{Dec})$ are trained under the objective in Eq.~\eqref{eq:ctf},  we freeze these components and shift the generative task to the learned hierarchical token space. 
This allows us to reformulate the data log-likelihood, $\log p_\theta(\mathbf{x})$, in terms of the corresponding tokens using a block-wise autoregressive factorization: $\log p_\phi\big(\mathbf{Q}_{1:n_L}\big) 
    = \sum_{\ell=1}^{L}
    \log p_\phi\big(
        \mathbf{Q}_{n_{\ell-1}+1:n_\ell}
        \,\big|\,
        \mathbf{Q}_{1:n_{\ell-1}}
    \big)$ 
where the conditional distributions are parameterized by~ $\phi$ and implemented using a VAR Transformer~\citep{tian_VAR_2024NeurIPS}. 
Unlike conventional autoregressive models that generate sequences via token-by-token, next-token prediction, our VAR Transformer operates at the level of granularity blocks. It predicts a group of tokens corresponding to the next temporal granularity level, conditioned on all preceding coarser-level tokens as a prefix.  
We train $p_\phi$ using the following block-wise autoregressive objective: 
\begin{equation}
    \mathcal{L}(\phi) = - \mathbb{E}_{\mathbf{Q}} \left[ \textstyle\sum_{\ell=1}^{L} \log p_\phi\big(
        \mathbf{Q}_{n_{\ell-1}+1:n_\ell}
        \,\big|\,
        \mathbf{Q}_{1:n_{\ell-1}} ) \right],
\end{equation}
where $\mathbf{Q}$ is sampled from the tokenized time series in the training set. By sequentially generating token blocks across granularity levels, the model learns to refine coarse time-series representations with progressively finer details. This sequential block generation realizes Eq.~\eqref{eq:Chain_Rule} in the token space, capturing the full refinement trajectory with only $M$ tokens -- \ie the number of tokens required to represent a finest-grained time series -- that enable efficient \gctsg.

\subsection{Granularity-Controllable Generation~(Inference)}\label{sec:gctsg_infer}
The key capability of \name is its ability to generate fine-grained time series from coarse inputs, such as low-fidelity series or user-provided sketches. Given a source granularity level $i$ and a target level $j$, \name performs \gctsg through the following three steps: 
\begin{enumerate}[label=\textcircled{\scriptsize\arabic*},leftmargin=1.2em,topsep=0pt, partopsep=0pt,itemsep=0pt]
    \item \textbf{Input Tokenization.} Since the input granularity is not known a priori, we first encode the input time series into a sequence of tokens $\mathbf{Q}$. We then determine its granularity level $i$ by evaluating the reconstruction error at each level~(detailed in~\cref{appendix:level_measurement}). The prefix $\mathbf{Q}_{1:n_i}$, corresponding to the inferred level $i$, is then extracted to serve as the initial conditioning for the VAR Transformer.\vspace{-0.1em}
    \item \textbf{Token Generation.} Using $\mathbf{Q}_{1:n_i}$ as the initial context, we generate tokens in a coarse-to-fine manner. For each successive level $\ell = i{+}1, \ldots, j$, we sample the corresponding token block: $\hat{\mathbf{Q}}_{n_{\ell-1}+1:n_\ell} \sim p_\phi\!\big(\cdot \big| \mathbf{Q}_{1:n_i}, \hat{\mathbf{Q}}_{n_i+1:n_{\ell-1}}\big)$ conditioning on both the original prefix and all previously generated tokens (with $\hat{\mathbf{Q}}_{n_i+1:n_{\ell-1}}\!=\!\varnothing$ for $\ell\!=\!i{+}1$). This iterative process continues until we obtain the complete token sequence with granularity level $j$.  \vspace{-0.1em}
    \item \textbf{Decoding.} The resulting token sequence is decoded back to the time-series representation at granularity level $j$, \ie $\hat{\mathbf{x}}^{(j)} = \texttt{DEC}\big(\mathbf{Q}_{1:n_i}\oplus\hat{\mathbf{Q}}_{n_{i}+1:n_j}\big)$.
\end{enumerate}
Generation from scratch is a special case in which no coarse input is provided. In this setting, the conditioning context reduces to the \texttt{[BOS]} token, from which we autoregressively generate $n_j$ tokens. Decoding the resulting tokens gives a time series at granularity level $j$. Similarly, generating all $n_L$ tokens becomes standard time-series generation. In both conditional and unconditional settings, class conditioning can be incorporated via AdaLN~\citep{peebles2023scalable}.

\section{Experiments }
\label{sec:experiments}

We provide the details of the datasets and experiments. Full implementation details, evaluation metrics, and ablation study can be found in Appendix \ref{app:Implementation Details} and \ref{app:Ablation}, respectively.

\define{Datasets \& Generation Tasks} We evaluate \name on two classification datasets (ECG5000 and ItalyPowerDemand) and two forecasting datasets~(Nasdaq and ETTh1). For standard generation, where time series are generated from scratch (\texttt{[BOS]} token), we measure the quality and distributional similarity between the synthetic and real data. In \gctsg, we generate $\hat{\mathbf{x}}^{(j)}$ from~$\mathbf{x}^{(i)}$~($j>i>0$), and evaluate (i) the diversity of the generated sequences $\hat{\mathbf{x}}^{(j)}$ and its alignment with $\mathbf{x}^{(j)}$, and (ii) structural consistency between $\mathbf{x}^{(i)}$ and $\hat{\mathbf{x}}^{(j)}$.

\define{Evaluation Metrics} In standard generation, we evaluate whether the generated time series aligns with real data. To assess this alignment, we measure (i) predictive score~\citep{yoon_timegan_2019NeurIPS}~-- \ie the downstream classification or forecasting performance in a Train-Synthetic-Test-Real (TSTR) setup~-- and (ii)~FID score~-- \ie distributional similarity between synthetic and real data, for which we adopt the FID evaluation module from DiffusionTS~\citep{yuandiffusion}. For \gctsg, we evaluate the ability to generate diverse fine-grained time-series $\hat{\mathbf{x}}^{(j)}$ from coarse inputs $\mathbf{x}^{(j)}$, while preserving the structure of the coarse series. As such, we use $\text{CRPS}_{\text{sum}}$ (Continuous Ranked Probability Score)~\citep{fan_mgtsd_2024ICLR}, a commonly used metric for probabilistic forecasting, to measure diversity and consistency. We also measure the C-Cons (coarse consistency) between~$\mathbf{x}^{(i)}$ and re-coarsened $\hat{\mathbf{x}}^{(j)}$ using RMSE.  

\define{Baseline Methods} We compare \name against state-of-the-art methods including RNNs trained with teacher forcing~(TF)~\citep{sutskever2011generating}, professor forcing~(PF)~\citep{lamb2016professor}, TSGDiff~\cite{shen_tsgdiff_AAAI2026}, TimeGAN~\citep{yoon_timegan_2019NeurIPS}, TimeVQVAE~\citep{Daesoo_TimeVQVAE_2023AISTATS}, DiffusionTS~\citep{yuandiffusion}, TOTEM~\citep{talukdertotem}, and ImagenTime~\cite{naiman_ImagenTime_NeurIPS2024}. TSGDiff and ImagenTime are evaluated only for standard generation, as extending them to GC-TSG requires substantial architectural modifications; details on baseline selection are provided in~\cref{appendix:baseline_selection}.


\define{Extension to Foundational Tokenizer} So far, we have assumed that the granularity level of a given time-series dataset corresponds to the finest granularity level (\ie $\mathbf{x}^{(L)} \ldefeq \mathbf{x}$). This assumption, however, may not hold in heterogeneous settings, such as foundation model training, where data naturally spans multiple temporal granularities, and uniformly treating all time series as finest-granular may be inappropriate. To address this gap, we extend \name by determining the granularity level using DFA, rather than uniformly treating all samples as level $L$. We then train the hierarchical tokenizer on UTSD~\citep{liutimer}, a multi-domain time-series collection designed for foundation models.  We evaluate the transferability of \name under this setup; more details are provided in~\cref{subsec:foundation}.

\vspace{-3mm}



\subsection{Standard Time-Series Generation}
\label{subsec:results_classification}
\begin{figure}[t]
\begin{center}
\center{\includegraphics[width=\columnwidth]{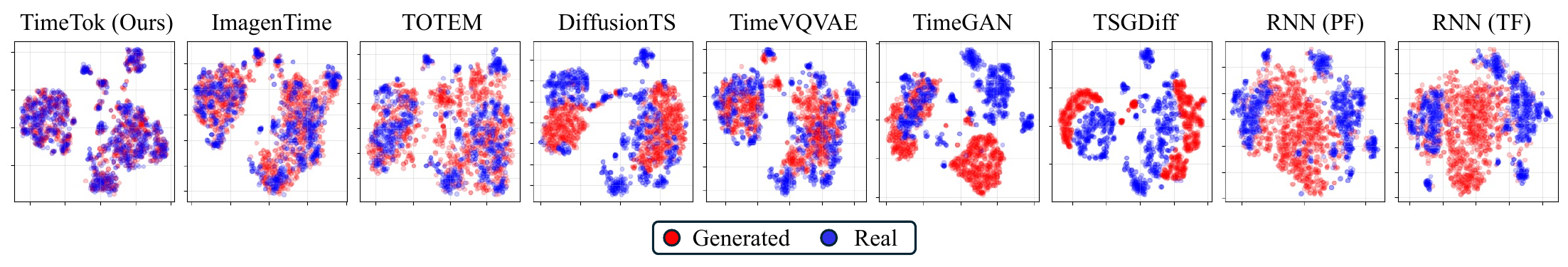}} \vspace{-1mm}
    \vspace{-15pt}
    \caption{\textbf{t-SNE visualization of ECG5000}. We observe a higher overlap between the synthetic and real data distribution for \name, while other methods show signs of mode collapse.} 
    \label{fig:tsne_main} 
    \vspace{-15pt}
\end{center}
\end{figure}

\define{Quantitative Performance} For the TSTR setup, we train a backbone model~(\ie InceptionTime~\citep{ismail2020inceptiontime} for classification, DLinear~\citep{zeng2023transformers} for forecasting) and report its predictive performance. We provide class conditioning to all models used in classification. As shown in~\cref{tab:results_standard}, \name achieves the best performance on ECG5000, ItalyPowerDemand, and Nasdaq, and competitive performance on ETTh1, in both predictive and FID metrics. Here, Oracle indicates the predictive performance when the prediction models are trained on the real data.

\define{Visualizations with t-SNE}
\cref{fig:tsne_main} shows t-SNE visualizations of the ECG5000 dataset. Synthetic data generated by \name~(\red{red}) strongly overlaps with the real data distribution (\blue{blue}), indicating high diversity and fidelity. In contrast, baseline methods show evidence of mode collapse or incomplete coverage of the true distribution.

\subsection{Granularity-Controllable Time-Series Generation}
\label{subsec:results_gc}
We evaluate \name on \gctsg with two related tasks: (i) \textit{conditional} generation, where given a coarse input, each model generates diverse and finer-grained time series at a specified target granularity while preserving the coarse structure, and (ii) \textit{unconditional} generation, where each model generates time series at a specified target granularity level from scratch. We provide the results in~\cref{tab:conditional_GC_TSG_main} and~\cref{tab:marginal_generation_main}, respectively. 
For fair comparison, we adapt the baselines to support granularity control. Since these methods cannot inherently generate time series at a specified granularity, we train them on datasets containing paired coarse- and fine-grained representations, allowing conditional generation from a coarse input to a specified target granularity. For unconditional generation, we replace the coarse input with a zero or \texttt{Null} embedding. See details in \cref{appendix:baseline}.

\begin{table*}[t]
\centering
\small
\setlength{\tabcolsep}{2pt} 
\sisetup{detect-weight=true,detect-inline-weight=math}
\caption{\small\define{Standard Generation} Generation performance on classification~(ECG5000, ItalyPowerDemand; ACC/F1) and forecasting~(Nasdaq, ETTh1; MSE) tasks, with FID reported across all datasets. }
\vspace{-5pt}
\label{tab:results_standard}
\resizebox{0.97\textwidth}{!}{%
\begin{tabular}{l|ccc|ccc|cc|cc}
\toprule
\multirow{2}{*}{Method} &
\multicolumn{3}{c}{ECG5000} &
\multicolumn{3}{c}{ItalyPowerDemand} &
\multicolumn{2}{c}{Nasdaq} &
\multicolumn{2}{c}{ETTh1} \\
\cmidrule(lr){2-4}
\cmidrule(lr){5-7}
\cmidrule(lr){8-9}
\cmidrule(l){10-11}
& FID \color{blue}$\downarrow$& ACC \color{red}$\uparrow$& F1 \color{red}$\uparrow$& FID \color{blue}$\downarrow$& ACC \color{red}$\uparrow$& F1 \color{red}$\uparrow$& FID \color{blue}$\downarrow$& MSE \color{blue}$\downarrow$& FID \color{blue}$\downarrow$& MSE \color{blue}$\downarrow$\\
\midrule
Oracle &
-- & 0.959\stdtiny{0.001} & 0.620\stdtiny{0.010} &
-- & 0.982\stdtiny{0.008} & 0.982\stdtiny{0.008} &
-- & 0.033\stdtiny{0.000} &
-- & 0.098\stdtiny{0.000} \\
\midrule
RNN(TF) &
0.220\stdtiny{0.047} & 0.895\stdtiny{0.014} & 0.383\stdtiny{0.024} &
0.125\stdtiny{0.034} & 0.903\stdtiny{0.027} & 0.903\stdtiny{0.027} &
0.462\stdtiny{0.083} & 0.046\stdtiny{0.000} &
1.195\stdtiny{0.040} & 0.104\stdtiny{0.000} \\
RNN(PF) &
0.076\stdtiny{0.028} & 0.884\stdtiny{0.013} & 0.398\stdtiny{0.035} &
0.101\stdtiny{0.019} & 0.952\stdtiny{0.020} & 0.952\stdtiny{0.021} &
1.070\stdtiny{0.119} & 0.054\stdtiny{0.001} &
0.730\stdtiny{0.015} & 0.112\stdtiny{0.001} \\
 TSGDiff& 0.102\stdtiny{0.033}& 0.854\stdtiny{0.006}& 0.467\stdtiny{0.047}& 0.088\stdtiny{0.011}& 0.931\stdtiny{0.027}& 0.931\stdtiny{0.028}& 0.786\stdtiny{0.085}& 0.070\stdtiny{0.000}& 1.185\stdtiny{0.150}&0.111\stdtiny{0.000}\\
TimeGAN &
0.034\stdtiny{0.002} & 0.815\stdtiny{0.042} & 0.348\stdtiny{0.018} &
0.026\stdtiny{0.008} & 0.958\stdtiny{0.003} & 0.958\stdtiny{0.003} &
0.316\stdtiny{0.050} & 0.061\stdtiny{0.001} &
1.068\stdtiny{0.243} & 0.104\stdtiny{0.001} \\
TimeVQVAE &
0.033\stdtiny{0.011} & 0.925\stdtiny{0.004} & 0.380\stdtiny{0.001} &
0.060\stdtiny{0.011} & 0.979\stdtiny{0.003} & 0.979\stdtiny{0.003} &
0.136\stdtiny{0.015} & 0.048\stdtiny{0.001} &
0.107\stdtiny{0.009} & 0.102\stdtiny{0.001} \\
DiffusionTS &
0.041\stdtiny{0.009} & 0.896\stdtiny{0.015} & 0.552\stdtiny{0.028} &
0.023\stdtiny{0.005} & 0.979\stdtiny{0.003} & 0.979\stdtiny{0.003} &
0.208\stdtiny{0.015} & 0.056\stdtiny{0.001} &
0.083\stdtiny{0.011} & 0.101\stdtiny{0.001}\\
TOTEM &
0.022\stdtiny{0.006} & 0.923\stdtiny{0.004} & 0.379\stdtiny{0.001} &
0.016\stdtiny{0.004} & 0.979\stdtiny{0.007} & 0.979\stdtiny{0.007} &
0.164\stdtiny{0.008} & 0.046\stdtiny{0.001} &
0.102\stdtiny{0.015} & 0.102\stdtiny{0.001} \\
 ImagenTime& 0.020\stdtiny{0.010}& 0.940\stdtiny{0.007}& 0.477\stdtiny{0.104}& 0.043\stdtiny{0.006}& 0.978\stdtiny{0.002}& 0.978\stdtiny{0.002}& 0.142\stdtiny{0.032}& 0.046\stdtiny{0.000}& 0.074\stdtiny{0.007}&\textbf{0.099\stdtiny{0.000}}\\
\midrule
\rowcolor[HTML]{F0F8F8} 
\textbf{\name}&
\textbf{0.004\stdtiny{0.001}} & \textbf{0.948\stdtiny{0.006}} & \textbf{0.564\stdtiny{0.047}} &
\textbf{0.007\stdtiny{0.001}} & \textbf{0.982\stdtiny{0.000}} & \textbf{0.982\stdtiny{0.000}} &
\textbf{0.091\stdtiny{0.017}} & \textbf{0.045\stdtiny{0.001}} &
\textbf{0.070\stdtiny{0.006}} & 0.103\stdtiny{0.000} \\
\bottomrule
\end{tabular}
}\vspace{-3mm}
\end{table*}

\define{Conditional Generation} We evaluate conditional generation using two metrics~(details in~\cref{app:Implementation Details}). First, we generate five time series for each coarse input, and measure the $\text{CRPS}_{\text{sum}}$ (lower is better) to evaluate the diversity and its preservation of global structure. \name consistently outperforms all baselines in both datasets. Second, we measure coarse-level consistency~(C-Cons) by re-coarsening the generated time series and computing the RMSE against the original coarse input. \name again achieves the best performance across all baselines.

\define{Unconditional Generation} Using the \texttt{[BOS]} token, \name generates time series at granularity levels 1 through 8, respectively; level 8 corresponds to standard generation. \cref{tab:marginal_generation_main}  reports FID averaged over all levels, showing that TimeTok achieves the best performance compared to baselines.


\define{Token-Space Refinement Matters}  To test whether \name's gains come from the coarse-to-fine factorization alone or from learning it in a structured token space, we compare against a raw-space AR Transformer that directly models the full refinement trajectory (\ie direct implementation of~Eq.~\eqref{eq:Chain_Rule}). As shown in~\cref{tab:architecture_ablation_conditional}, TimeTok consistently improves both distributional accuracy and coarse consistency, indicating that GC-TSG requires not only a coarse-to-fine objective, but a representation in which coarse-to-fine refinement is compact, ordered, and generative.  


\begin{table*}[!t]
\centering
\scriptsize

\begin{minipage}[t]{0.57\textwidth}
\centering
\setlength{\tabcolsep}{3.0pt}
\caption{\small \define{Conditional Generation}
Conditional GC-TSG performance averaged over all granularity pairs~(Avg, $i\rightarrow j$, $0<i< j$). C-Cons denotes coarse-level consistency. Full
results of \name are in~\cref{app:Fullresult}.}
\label{tab:conditional_GC_TSG_main}
\resizebox{\linewidth}{!}{%
\begin{tabular}{l |cc |cc}
\toprule
&\multicolumn{2}{c}{\textbf{Classification (ECG5000)}}&
\multicolumn{2}{c}{\textbf{Forecasting (Nasdaq)}}\\
\cmidrule(lr){2-3}\cmidrule(lr){4-5}

\textbf{Method}
& \textbf{$\text{CRPS}_{\text{sum}}$} \color{blue}$\downarrow$
& \textbf{C-Cons} \color{blue}$\downarrow$
& \textbf{$\text{CRPS}_{\text{sum}}$} \color{blue}$\downarrow$
& \textbf{C-Cons} \color{blue}$\downarrow$ \\
\midrule

RNN(TF) &
5.602\stdtiny{1.869} & 0.059\stdtiny{0.020} & 86.72\stdtiny{52.09} & 0.483\stdtiny{0.295}\\

RNN(PF) &
3.797\stdtiny{1.067} & 0.034\stdtiny{0.011} & 48.20\stdtiny{22.14} & 0.370\stdtiny{0.215}\\

TimeGAN &
1.737\stdtiny{0.737} & 0.011\stdtiny{0.005} & 39.11\stdtiny{22.14} & 0.196\stdtiny{0.105} \\

TimeVQVAE &
3.693\stdtiny{0.858} & 0.044\stdtiny{0.013} & 35.32\stdtiny{34.30} & 0.282\stdtiny{0.258} \\

DiffusionTS &
2.672\stdtiny{1.551} & 0.031\stdtiny{0.019} & 101.5\stdtiny{63.26} & 0.771\stdtiny{0.538} \\

TOTEM &
2.639\stdtiny{1.286} & 0.035\stdtiny{0.019} & 35.97\stdtiny{22.54} & 0.294\stdtiny{0.247} \\

\midrule
\rowcolor[HTML]{F0F8F8} 
\textbf{\name}&
\textbf{1.587\stdtiny{0.867}} & \textbf{0.010\stdtiny{0.006}} &
\textbf{15.87\stdtiny{12.48}}  & \textbf{0.096\stdtiny{0.090}} \\
\bottomrule 
\end{tabular}
}
\end{minipage}
\hspace{0.02\textwidth}
\begin{minipage}[t]{0.38\textwidth}
\centering
\setlength{\tabcolsep}{3.0pt}
\caption{\small\define{Unconditional Generation}
Unconditional generation to each granularity level, averaged over all levels. Full results are provided in~\cref{app:Fullresult}.}
\label{tab:marginal_generation_main}

{\renewcommand{\arraystretch}{0.887}
\resizebox{\linewidth}{!}{%
\begin{tabular}{l |c |c}
\toprule
& \textbf{ECG5000}& \textbf{Nasdaq}\\
\cmidrule(lr){2-2}\cmidrule(lr){3-3}

\textbf{Method}
& \textbf{FID} \color{blue}$\downarrow$
& \textbf{FID} \color{blue}$\downarrow$ \\
\midrule

RNN(TF)      & 0.884\stdtiny{0.316}& 8.339\stdtiny{0.856}\\
RNN(PF)      & 0.251\stdtiny{0.047}& 4.199\stdtiny{0.351}\\
TimeGAN      & 0.788\stdtiny{0.194}& 9.568\stdtiny{0.658}\\
TimeVQVAE    & 0.029\stdtiny{0.006}& 1.039\stdtiny{0.072}\\
DiffusionTS  & 0.020\stdtiny{0.005}& 0.432\stdtiny{0.048}\\
TOTEM        & 0.029\stdtiny{0.007}& 0.724\stdtiny{0.049}\\

\midrule
\rowcolor[HTML]{F0F8F8} 
\textbf{\name}& \textbf{0.002\stdtiny{0.001}}& \textbf{0.103\stdtiny{0.014}}\\
\bottomrule
\end{tabular}
}
}

\end{minipage}

\vspace{-3mm}
\end{table*}

\define{Bridging Temporal Granularity Mismatch} We have performed an additional experiment to verify that coarse-to-fine generated time series are indeed helpful in the downstream tasks (classification), particularly under temporal granularity mismatch between the model’s training data and test samples. Specifically, we evaluate classifiers trained on fine-grained ECG signals under four test settings: classifiers tested on (1) coarse time series (obtained by downsampling followed by linear interpolation), (2) coarse time series with Gaussian noise added, (3) refined time series by \name, and (4) original fine-grained time series (Oracle). This setup reflects the realistic scenario discussed in the Introduction, where consumer-grade devices provide only low-resolution ECG signals, whereas hospital-trained models are trained using fine-grained time series. \cref{tab:Downstream_Performance} demonstrates that \name can effectively adjust temporal granularity without deviating from the underlying data distribution.

\begin{table*}[t]
\centering

\begin{minipage}[t]{0.36\textwidth}
\centering
\caption{\small\textbf{Bridging Temporal Granularity Mismatch.} Classification on ECG5000 with different granularity.}
\label{tab:Downstream_Performance}\vspace{-2.mm}

{\renewcommand{\arraystretch}{1.25}
\resizebox{\linewidth}{!}{%
\begin{tabular}{lc}
\toprule
\textbf{ECG5000} & \textbf{ACC} \color{red}$\uparrow$\\
\midrule
Oracle & 0.959\stdtiny{0.001}\\
\midrule
Coarse Time Series & 0.314\stdtiny{0.046}\\
Coarse Time Series (+Noise) & 0.314\stdtiny{0.045}\\
\rowcolor[HTML]{F0F8F8} 
\textbf{Refined Time Series (\name)}& \textbf{0.918\stdtiny{0.006}}\\
\bottomrule
\end{tabular}
}
}
\end{minipage}%
\hfill
\begin{minipage}[t]{0.61\textwidth}
\centering
\scriptsize
\caption{\small\textbf{Foundation tokenizer evaluation} on ECG5000 and Nasdaq datasets. 
Metrics: FID (lower is better), ACC, F1, and MSE.}
\label{tab:results_FM} \vspace{-2mm}
{\renewcommand{\arraystretch}{0.91}
\resizebox{\linewidth}{!}{%
\begin{tabular}{l|ccc cc}
\toprule
& \multicolumn{3}{c}{\textbf{ECG5000}} 
& \multicolumn{2}{c}{\textbf{Nasdaq}} \\
\cmidrule(lr){2-4}\cmidrule(lr){5-6}
\textbf{Method} 
& FID \color{blue}$\downarrow$
& ACC \color{red}$\uparrow$
& F1 \color{red}$\uparrow$
& FID \color{blue}$\downarrow$
& MSE \color{blue}$\downarrow$\\
\midrule
TimeVQVAE & 0.367 & 0.924 & 0.450 & 0.082 & 0.047 \\
TOTEM     & 0.035 & 0.929 & 0.394 & 0.068 & 0.045 \\
\midrule
\rowcolor[HTML]{F0F8F8} 
\textbf{\name (w/o DFA)}& 0.042 & 0.944 & 0.583 & 0.111 & 0.044 \\
\rowcolor[HTML]{F0F8F8} 
\textbf{\name (w/ DFA)}& \textbf{0.010} & \textbf{0.951} & \textbf{0.668} & \textbf{0.046} & \textbf{0.042} \\
\bottomrule
\end{tabular}
}
}
\end{minipage}

\vspace{-0.3mm}
\end{table*}



\subsection{Extension to Foundational Time-series Tokenization}
\label{subsec:foundation}
In the foundational setting, we train \name on UTSD \citep{liutimer}. We apply the same $\mathcal{R}_{\ell}$'s used in the individual setup, but determine the absolute granularity level for each time series using the DFA scaling exponent, which is discretized into levels via uniform binning. Based on these estimated granularities, we first train the hierarchical tokenizer on UTSD and then train the VAR transformer on downstream tasks using this foundational tokenizer.  Details are provided in~\cref{appendix:Extension_FM} 

In~\cref{tab:results_FM}, we evaluate the standard generation performance of our foundational tokenizer using two variants of \name: (i)~w/o DFA, using the individual setup, and (ii)~w/ DFA, using the DFA-based granularity levels. For fair comparison, we also train TimeVQVAE and TOTEM on UTSD. Both variants outperform TimeVQVAE, with the DFA-based variant showing a substantial performance gain. We attribute this improvement to training on a large, heterogeneous dataset with diverse temporal granularities, which helps the tokenizer better capture granularity-specific structures. These results underscore \name's promise as a foundational tokenizer for time series.
\vspace{-5pt}

\subsection{Qualitative and Additional Analysis}
\vspace{-3pt}
\begin{figure}[!t]
\begin{center}
\vspace{-5pt}
\center{\includegraphics[width=0.93\columnwidth]{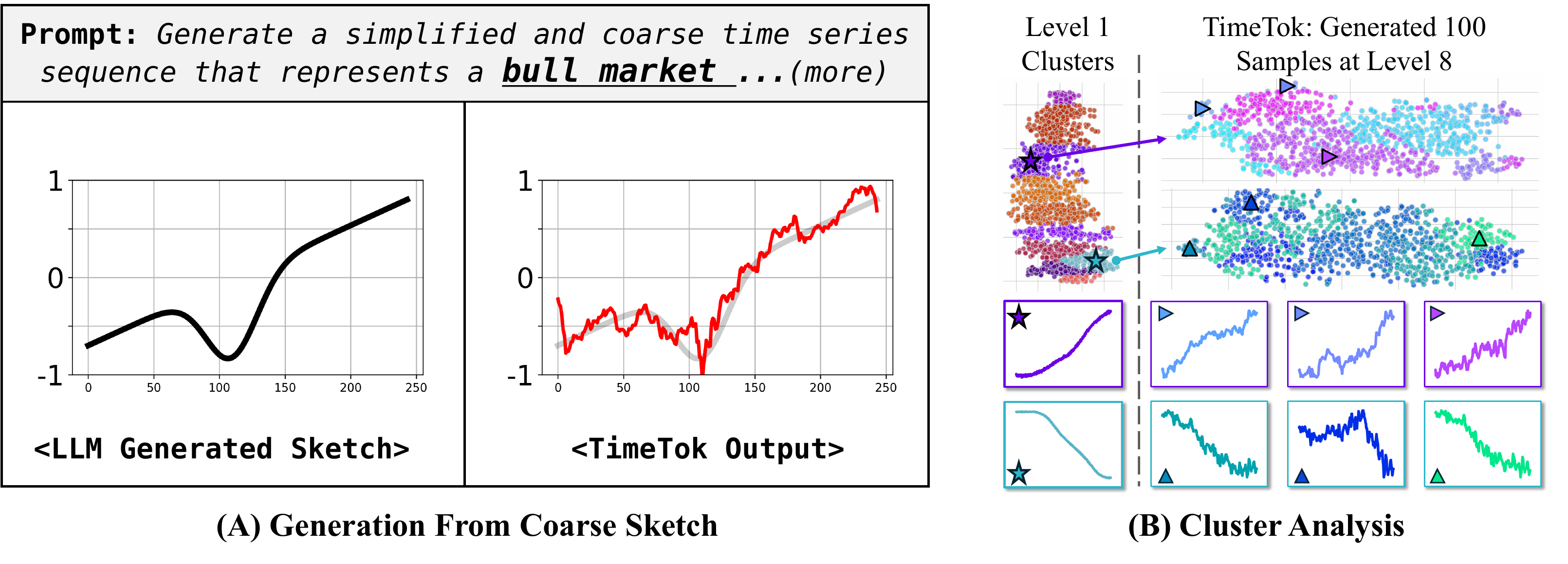}}
\vspace{-5pt}
    \caption{\small \define{(A) Generation From Coarse Sketch} We visualize fine-grained time series generated from coarse sketches provided by the user. \define{(B) Cluster Analysis} Using the token in Level 1 as a condition, we generate samples from Level 8. We observe high diversity, showcasing \name's performance as a generative model.} 
    \label{fig:bull_market} 
    \vspace{-10pt}
\end{center}
\end{figure}

\label{subsec:qualitative_anal}
\define{User-provided Coarse Sketch} We generate a coarse input via an LLM~(\eg \emph{an increasing stock trend} in \cref{fig:bull_market}-A) and use \name to transform this coarse sketch into a fine-grained, realistic time series. See full results in~\cref{appendix:use_case}.

\define{Clustering Analysis by Granularity} We perform a clustering analysis of the generated time series based on their granularity level~(\cref{fig:bull_market}-B). As illustrated in the figure, the model generates diverse samples enriched with fine-grained details, while preserving the coarse trend in Level 1.

\define{Token-usage Analysis}
We analyze the token-usage pattern of \name on Nasdaq in~\cref{appendix:token_usage}. 


\define{\gctsg} We visualize the coarse-to-fine refinement process in \gctsg (\cref{appendix:gctsg_full}).  

\define{Information-Reducing Operators}  We evaluate the robustness of our method using a test set constructed with various information-reducing operators (\cref{appendix:info_reducing}).

\define{Ablation} We conduct further ablation studies on key design components of \name to assess their impact and explore alternative design choices (\cref{app:Ablation}).

\section{Conclusion}
In this work, we proposed~\name, a novel time-series generation model capable of generating time series at any target granularities, whether conditioned on coarse-grained inputs or generated entirely from scratch. This flexibility is made possible with our proposed hierarchical tokenization strategy, where simply controlling the number of tokens for decoding controls the granularity level of the output -- enabling truly user-driven generation of time series. We envision \name as a promising generative model that can enable the acquisition of diverse multi-granularity data needed for training time-series foundation models, while also solving a wide range of practical, real-world problems.

\paragraph{Acknowledgements}
We thank the NeurIPS reviewers for their valuable comments and suggestions. This work was supported by the National Research Foundation of Korea (NRF) grant funded by the Korean government (MSIT) (No. RS-2024-00358602); the Institute of Information \& Communications Technology Planning \& Evaluation (IITP) grants funded by the Korean government (MSIT), the Artificial Intelligence Graduate School Program (No. RS-2019-II190079, Korea University) and the AI Research Hub Project (No. RS-2024-00457882); and the National Institute for Mathematical Sciences (NIMS) grant funded by the Korean government (MSIT) (No. S26120000). 
There are no competing interests to declare.

\bibliographystyle{unsrt}
\bibliography{Main_paper}

\newpage
\appendix
\clearpage
{\Huge\bfseries Appendix\par}
\vspace{1em}
{\Large\bfseries Table of Contents\par}
\vspace{0.3em}
\hrule
\vspace{0.8em}

\noindent\textbf{A\quad EXPERIMENT DETAILS}
\hfill\textbf{\pageref{app:Implementation Details}}\\[0.2em]
\hspace*{1.5em} A.1\quad Gaussian kernel smoothing
\hfill\pageref{Appendix-subsec:Gaussian-kernel-smoothing}\\[0.2em]
\hspace*{1.5em} A.2\quad DATASETS
\hfill\pageref{app:dataset_details}\\[0.2em]
\hspace*{1.5em} A.3\quad Level Measurement
\hfill\pageref{appendix:level_measurement}\\[0.2em]
\hspace*{1.5em} A.4\quad Extension to Foundational Tokenizer
\hfill\pageref{appendix:Extension_FM}\\[0.2em]
\hspace*{1.5em} A.5\quad Evaluation Metric
\hfill\pageref{appendix:eval_metric}\\[0.2em]
\hspace*{1.5em} A.6\quad Baseline Selection for \gctsg
\hfill\pageref{appendix:baseline_selection}\\[0.2em]
\hspace*{1.5em} A.7\quad Baseline Implementation for \gctsg
\hfill\pageref{appendix:baseline}\\[0.4em]

\noindent\textbf{B\quad Limitations and Broader Impacts}
\hfill\textbf{\pageref{app:limitations_broaderImpacts}}\\[0.4em]

\noindent\textbf{C\quad Related Works}
\hfill\textbf{\pageref{app:related_work}}\\[0.2em]
\hspace*{1.5em} C.1\quad Time-Series Generation
\hfill\pageref{app:related_TSG}\\[0.2em]
\hspace*{1.5em} C.2\quad Multi-Granular and Discrete Time-Series Representation Learning
\hfill\pageref{app:related_MG}\\[0.2em]
\hspace*{1.5em} C.3\quad Multi-Scale Token Generation in Computer Vision
\hfill\pageref{app:related_MS}\\[0.2em]
\hspace*{1.5em} C.4\quad Difference with Temporal Super-Resolution
\hfill\pageref{app:related_TSR}\\[0.2em]
\hspace*{1.5em} C.5\quad Relation to Equivariance Learning
\hfill\pageref{app:equivariance}\\[0.4em]

\noindent\textbf{D\quad ADDITIONAL EXPERIMENTAL RESULTS}
\hfill\textbf{\pageref{app:Fullresult}}\\[0.2em]
\hspace*{1.5em} D.1\quad Compute Resource and Experiment Environment
\hfill\pageref{app:compute_resource}\\[0.2em]
\hspace*{1.5em} D.2\quad Full Results
\hfill\pageref{app:full_results_table}\\[0.2em]
\hspace*{1.5em} D.3\quad Ablation Study
\hfill\pageref{app:Ablation}\\[0.2em]
\hspace*{1.5em} D.4\quad Granularity Monotonicity and Smoothing Supervision
\hfill\pageref{app:granularity_monotonicity}\\[0.2em]
\hspace*{1.5em} D.5\quad Long-Sequence Generation and Model Size
\hfill\pageref{subsec:long_sequence}\\[0.2em]
\hspace*{1.5em} D.6\quad Model Parameter Count
\hfill\pageref{app:parameter_count}\\[0.4em]

\noindent\textbf{E\quad Coarse To Fine Reconstruction}
\hfill\textbf{\pageref{appendix:full_coarse_2_fine}}\\[0.4em]

\noindent\textbf{F\quad \gctsg Generation}
\hfill\textbf{\pageref{appendix:gctsg_full}}\\[0.4em]

\noindent\textbf{G\quad Other Information Reducing Operations}
\hfill\textbf{\pageref{appendix:info_reducing}}\\[0.4em]

\noindent\textbf{H\quad Use Case Analysis}
\hfill\textbf{\pageref{appendix:use_case}}\\[0.4em]

\noindent\textbf{I\quad The Use of LLMs in this work}
\hfill\textbf{\pageref{app:use_of_llms}}\\[0.4em]

\noindent\textbf{J\quad Token Usage Analysis on the Nasdaq Dataset}
\hfill\textbf{\pageref{appendix:token_usage}}\\[0.4em]

\noindent\textbf{K\quad Robustness Between Evaluators}
\hfill\textbf{\pageref{appendix:Robustness}}\\[0.4em]

\hrule
\clearpage

\section{EXPERIMENT DETAILS}
\label{app:Implementation Details}

\begin{figure}[!ht]
    \centering
    \includegraphics[width=\linewidth]{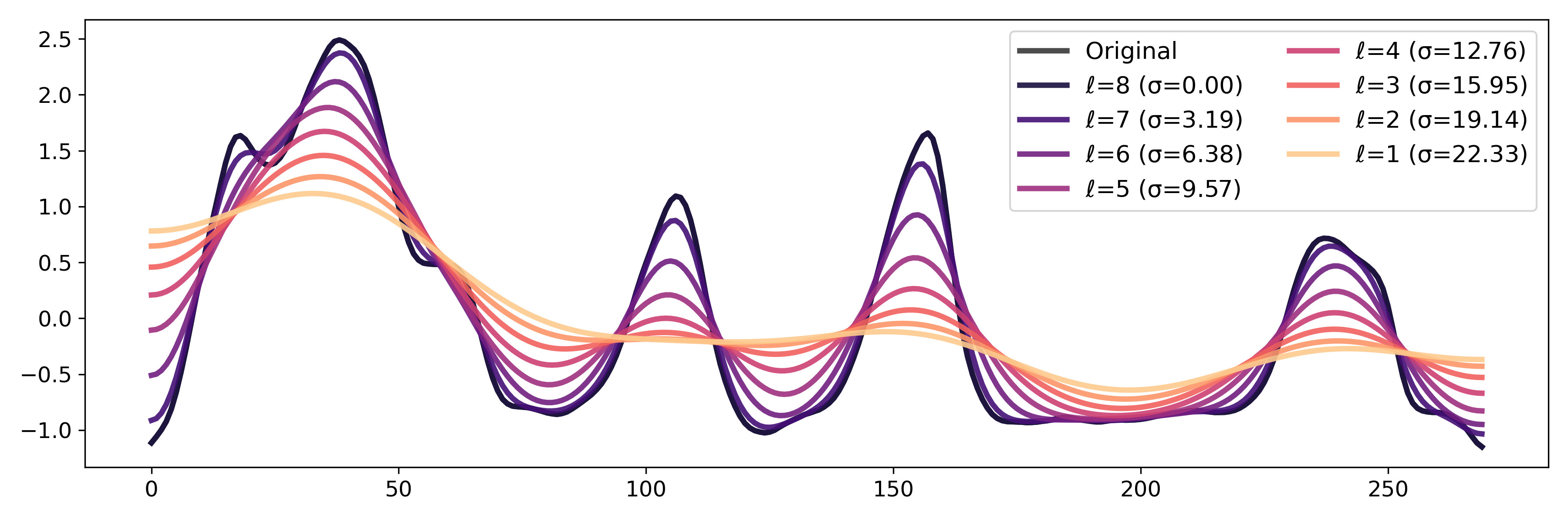}
    \caption{
        Example of Gaussian kernel smoothing with varying $\ell$ values. The original signal (\textit{Original}) is shown together with smoothed signals for different $\ell$. Note that $\ell=8$ is identical to the original signal.
    }
    \label{fig:smoothing}
\end{figure} 

\subsection{Gaussian kernel smoothing}
\label{Appendix-subsec:Gaussian-kernel-smoothing}

Target signal $\mathbf{x}^{(\ell)}$ is constructed by smoothing the original sequence $\mathbf{x}^{(L)} \ldefeq \mathbf{x}$ with a Gaussian kernel, $\mathbf{x}^{(\ell)} = g_{\sigma(\ell)} * \mathbf{x}$, where $g_{\sigma(\ell)}$ denotes a discrete normalized Gaussian kernel, \ie $\sum_i g_\sigma[i] = 1$. Specifically, the kernel weights are given by $g_{\sigma(\ell)}[i] \propto \exp\!\left(-\tfrac{i^2}{2\cdot {\sigma(\ell)}^2}\right)$ and truncated to the window $|i| \le 3\cdot {\sigma(\ell)}$. The bandwidth $\sigma(\ell)$ is defined as a function of the level $\ell$:
\[
\sigma(\ell)
= \sigma_{\max} \cdot
\left(
\frac{L - \ell}{L - 1}
\right)
\]
where we set $\sigma_{\max} = T/12$ in practice, which results in an effective maximum kernel support of approximately $6\cdot \sigma_{\max} = T/2$ due to truncation at $|i| \le 3 \cdot \sigma(\ell)$. This design assigns stronger smoothing to smaller $\ell$, yielding coarse representations dominated by long-range structure, while gradually reducing the smoothing as $\ell$ increases so that higher levels progressively recover fine-grained details. Since Gaussian kernel smoothing is not uniquely invertible, recovering fine-scale details requires learning progressive refinement across higher granularity levels.

\subsection{DATASETS}
\label{app:dataset_details}

\begin{table}[ht]
\centering
\caption{\define{Dataset Statistics} For each dataset, we report the number of samples in the original train/test splits, the total number of samples, the number of classes, and the length of the time series. We use the first channel of ETTh1. For the UTSD, we report the actual number of samples used to train \name.}
\label{tab:dataset_statistics}
\resizebox{\textwidth}{!}{%
\begin{tabular}{clrrrrrrr}
\toprule
\textbf{ID} & \textbf{Dataset Name} & \textbf{\# Train} & \textbf{\# Test} & \textbf{\# Total} & \textbf{\# Classes} & \textbf{Input Len} & \textbf{Pred Len} & \textbf{TS Len} \\
\midrule
1& ECG5000                     & 500  & 4500  & 5000   & 5  & -- & -- & 140    \\
2& ItalyPowerDemand            & 67   & 1029  & 1096   & 2  & -- & -- & 24     \\
3& Nasdaq& 5320 & 1520 & 6840 & --& 196& 48& 244\\
4& ETTh1& 4900  & 1400 & 6300 & --& 96& 48& 144\\
\midrule
5& UTSD & 2,954,239  & - & 2,954,239 & --& -& -& 256\\
\bottomrule
\end{tabular}%
}
\end{table}

The ECG5000 and ItalyPowerDemand are datasets sourced from the UCR Time Series Classification Archive~\cite{dau_ucrdataset_2019JAS}. For the UCR datasets, we follow the train-validation-test split of 60\%-20\%-20\% as in~\citep{liu_softshape_ICML2025}. ETTh1 is sourced from \url{https://github.com/thuml/autoformer}. UTSD can be accessed from \url{https://huggingface.co/datasets/thuml/UTSD}. Nasdaq dataset will be made publicly available.

\newpage
\subsection{Level Measurement}
\label{appendix:level_measurement}

Given a time series with unknown granularity, \name must identify the appropriate granularity level to perform \gctsg. While users can manually specify the level based on prior knowledge, such information may not always be available. To address this, we implement an automatic level measurement based on the bias-variance tradeoff principle~\citep{bishop2006pattern}:

\begin{enumerate}[label=\textcircled{\scriptsize\arabic*},leftmargin=1.2em,topsep=0pt, partopsep=0pt,itemsep=0pt]
    \item \define{Time-Series Tokenization} We tokenize the input time series at the finest granularity level, retaining all tokens.\vspace{-0.1em}
    
    \item \define{Distance Measurement} We decode the tokens at each granularity level $\ell \in \{1, \ldots, 8\}$ obtaining time-series at each granularity level. We then compute the L2 distance between the input time series and each reconstructed time series.\vspace{-0.1em}
    
    \item \define{Complexity-Simplicity Tradeoff} Our objective is to balance token efficiency with reconstruction distance. Rather than simply selecting the level with the minimum distance, we identify the level that yields the highest relative improvement~(\ie the steepest decrease in distance relative to the previous level). This approach selects a level that captures the coarse information while avoiding unnecessary complexity.\vspace{-0.1em}
\end{enumerate}

The use of our level-measurement function was a design choice, where other measurement functions~(\eg Knee-detection) can be used for level detection.

\paragraph{Detection accuracy and impact on generation.}
On ECG5000, our level detector achieves an exact-match accuracy of
47.5\% and an accuracy of 73.4\% within $\pm 1$ level.
Errors are concentrated at coarse levels, where neighboring
granularities are similar: the mean squared difference between
adjacent-level signals increases from 0.0038 (levels 1--2) to
0.0472 (levels 7--8).
Despite imperfect level detection, its impact on generation is limited
in this evaluation. Using detected levels yields a
$\mathrm{CRPS}_{\mathrm{sum}}$ of 1.587, compared with 1.509 using
ground-truth levels, and both outperform TimeGAN (1.737).
These results suggest that confusion between similar neighboring
levels has a relatively small effect on conditional generation.

\begin{table}[!ht]
\centering
\scriptsize
\setlength{\tabcolsep}{2.0pt}
\renewcommand{\arraystretch}{0.9}

\begin{tabular}{l|c}
\toprule
\small
\textbf{Method} & \textbf{$\text{CRPS}_{\text{sum}}$} \color{blue}$\downarrow$ \\
\midrule
TimeGAN (Best Baseline) &
1.737\stdtiny{0.737} \\
\midrule
\textbf{\name} with Level Detect (Ours)  &
1.587\stdtiny{0.867} \\

\name with GT-Level &
\textbf{1.509\stdtiny{0.883}} \\
\bottomrule
\end{tabular}
\vspace{5pt}
\caption{\define{Comparison Between Level Measurements}
To further evaluate the sensitivity of our method to the granularity level measurement performed by the proposed measurement module, we additionally conduct the GC-TSG generation task on the ECG5000 dataset using the ground-truth (GT) granularity level. TimeTok with our detection (1.587) performs very close to the oracle setting with GT levels (1.509), with a gap of only 0.078. Both settings substantially outperform the best baseline (TimeGAN, 1.737), confirming that our level measurement module is robust and introduces negligible performance degradation.}
\label{apptab:conditional_GC_TSG_main}
\end{table} \vspace{-3mm}

\subsection{Extension to Foundational Tokenizer}
\label{appendix:Extension_FM}

For the sake of developing a simple and accessible methodology in the single-dataset setting, we have so far assumed that the granularity level of a given time-series dataset corresponds to the finest granularity level (\ie $\mathbf{x}^{(L)} \ldefeq \mathbf{x}$). This assumption, however, may not hold in heterogeneous settings, such as foundation model training, where data naturally spans multiple temporal granularities, and uniformly treating all time series as finest-granular may be inappropriate. To accommodate such heterogeneous settings, we further extend our training methodology to enable learning from time-series data observed at different temporal granularities.

The key difference is that, since the identity $\mathbf{x}^{(L)} \ldefeq \mathbf{x}$ no longer holds, the granularity levels can no longer be defined solely through the information-reducing operators $\mathcal{R}_\ell$. Accordingly, instead of assigning levels based on the strength of the information-reducing operators, we use detrended fluctuation analysis (DFA) exponent~\citep{peng_DFA_1995AIP} as an absolute quantitative measure of temporal fluctuation to determine the granularity level of each observed time series. Specifically, we define the granularity levels according to the following procedure.

\begin{enumerate}[label=\textcircled{\scriptsize\arabic*},leftmargin=1.2em,topsep=0pt, partopsep=0pt,itemsep=0pt]
    \item \define{Constructing multi-granular representations} For each observed time series $\mathbf{x}$, we apply the information-reducing operators across all levels to construct representations with diverse degrees of temporal granularity.
    \item \define{Computing the DFA exponent} For each resulting representation, we compute the DFA exponent as an absolute quantitative measure of its temporal fluctuation.
    \item \define{Defining level thresholds} To determine the DFA exponent range associated with each granularity level, we partition the resulting DFA exponents into $L$ uniform bins. Each candidate representation is then assigned to the granularity level corresponding to the bin that contains its DFA exponent.
\end{enumerate}
During training, each multi-granular representation constructed in Step~\textcircled{\scriptsize 1} is used as input. Given its assigned granularity level, the decoder is conditioned only on the corresponding number of leading tokens and trained to reconstruct the input at that level.

Under this DFA-based training procedure, the granularity level of an input time series is determined by the bin containing its DFA exponent, without relying on the reconstruction-based level measurement in~\cref{appendix:level_measurement}. The experiments in \cref{tab:results_FM} show that the proposed extension to a foundational tokenizer effectively improves performance in heterogeneous settings, by leveraging DFA-based granularity assignment (\ie w/ DFA).

\subsection{Evaluation Metric}
\label{appendix:eval_metric}

\paragraph{CRPS:} For conditional \gctsg, our goal is to evaluate how accurately the model captures the conditional distribution $p(\mathbf{x}^{(j)} \mid \mathbf{x}^{(i)})$, where $\mathbf{x}^{(i)}$ denotes coarse time series and $\mathbf{x}^{(j)}$ the corresponding target sequence. Given a condition $\mathbf{x}^{(i)}$, the model generates $K$ samples $\{\hat{\mathbf{x}}^{(j,k)}\}_{k=1}^K$ (with $K{=}5$ in all experiments) intended to represent the conditional predictive distribution of $\mathbf{x}^{(j)}$. We evaluate the conditional predictive distribution pointwise in time using the Continuous Ranked Probability Score (CRPS), a strictly proper scoring rule for univariate predictive distributions. 

At each time step $t$, CRPS is estimated from the generated samples $\{\hat{x}_t^{(j,k)}\}_{k=1}^K$ and the ground-truth target value $x_t^{(j)}$ as
\vspace{-2mm}
\[ \widehat{\mathrm{CRPS}}(F_t, x_t^{(j)})
=
\frac{1}{K}\sum_{k=1}^K \lvert \hat{x}_t^{(j,k)} - x_t^{(j)} \rvert
-
\frac{1}{2K^2}\sum_{k=1}^K\sum_{k'=1}^K
\lvert \hat{x}_t^{(j,k)} - \hat{x}_t^{(j,k')} \rvert .\]
To assess the quality of the entire conditional trajectory, we aggregate the pointwise scores over the temporal
horizon and define
\vspace{-2mm}
\[\mathrm{CRPS}_{\text{sum}}
=
\sum_{t=1}^{T} \widehat{\mathrm{CRPS}}(F_t, x_t^{(j)}).\]
Following prior work on time-series evaluation, we report CRPS$_{\text{sum}}$ averaged over the test set~\citep{fan_mgtsd_2024ICLR}. A lower CRPS$_{\text{sum}}$ indicates that the generated samples more accurately represent the conditional distribution $p(\mathbf{x}^{(j)} \mid \mathbf{x}^{(i)})$, both in terms of accuracy and uncertainty calibration.

\paragraph{Coarse consistency:} In addition to CRPS, we measure \emph{coarse consistency}, which evaluates whether generated target sequences remain consistent with the provided coarse condition. Recall that information-reducing operator is defined as $\mathbf{x}^{(i)} = \mathcal{R}_i(\mathbf{x}^{(L)})$, \ie $\mathcal{R}_i$ is explicitly specified as a mapping from the finest-grained observations $\mathbf{x}^{(L)}$ to $\mathbf{x}^{(i)}$. Therefore, $\mathcal{R}_i$ is not directly applicable to an arbitrary intermediate-level signal $\mathbf{x}^{(j)}$. To define a consistency metric for a generated sample $\hat{\mathbf{x}}^{(j)}$, we instead require an operator that maps level $j$ to level $i$ (with $i<j$), which we denote by $\mathcal{R}_{j\rightarrow i}$. We then recoarsify each generated target sample via $\mathcal{R}_{j\rightarrow i}(\hat{\mathbf{x}}^{(j)})$ and compute the RMSE against the ground-truth condition $\mathbf{x}^{(i)}$:
\[\mathrm{C\text{-}cons}
=
\mathrm{RMSE}\!\left(\mathcal{R}_{j\rightarrow i}(\hat{\mathbf{x}}^{(j)}), \mathbf{x}^{(i)}\right),\]
averaged over generated samples and test pairs. This metric is well-defined when $\mathcal{R}$ is instantiated as Gaussian smoothing. In this case, a level is parameterized by the Gaussian standard deviation, and Gaussian filters are closed under composition: applying Gaussians with $\sigma_1$ and $\sigma_2$ sequentially is equivalent to a single Gaussian with $\sqrt{\sigma_1^2+\sigma_2^2}$. Hence, given the known smoothing parameters for levels $i$ and $j$, we can construct $\mathcal{R}_{j\rightarrow i}$ by applying an additional Gaussian whose variance matches the difference, \ie $\sigma_{j\rightarrow i}^2 = \sigma_i^2 - \sigma_j^2$. For other information-reducing operators, an unambiguous mapping $\mathcal{R}_{j\rightarrow i}$ is generally unavailable, so C-cons is reported only for Gaussian smoothing and omitted otherwise.

\subsection{Baseline Selection for \gctsg}
\label{appendix:baseline_selection}
In selecting baselines, we applied the following criteria to ensure a fair and reproducible comparison:
\begin{enumerate}[label=\textcircled{\scriptsize\arabic*},leftmargin=1.2em,topsep=0pt, partopsep=0pt,itemsep=0pt]
    \item Official code is publicly available.\vspace{-0.1em}
    
    \item The method addresses \textbf{unconditional} (and preferably also \textbf{class-conditional}) time-series generation, excluding methods developed for classification or forecasting~\cite{fan_Mgtsd_ICLR2024}.\vspace{-0.1em}
    
    \item Applicable to \textbf{univariate} time series, i.e., not methods developed specifically for \textbf{multivariate} setting~\cite{li_PaDTS_AAAI2025}.\vspace{-0.1em}

    \item Designed for \textbf{general time-series generation}, rather than only for a specialized format (e.g., tabular-specific settings~\cite{garuti_TabDiT_ICLR2025}).\vspace{-0.1em}

    \item Not restricted to a highly specific scenario such as \textbf{data-scarcity-oriented}~\cite{gonen_ImagenFew_NeurIPS2025} or \textbf{multi-domain generation}~\cite{huang_timedp_AAAI2025}.\vspace{-0.1em}
\end{enumerate}

Critically, a fair baseline must be applicable to \textbf{\gctsg} without introducing unfair architectural advantages. Recent methods such as ImagenTime and TSGDiff are strong performers on standard generation, but applying them to GC-TSG requires \textbf{substantial architectural modifications.} That is, \textbf{ImagenTime} operates in a Fourier-domain where controlling temporal granularity levels would itself make an entirely new line of research, while \textbf{TSGDiff} further converts the Fourier-domain representation into a graph-structured form, adding another layer of complexity that makes granularity-controllable generation fundamentally incompatible with its core design. Such modifications go far beyond minor adaptations and can potentially make the comparison less rigorous. In contrast, ToTem and the other selected baselines could be extended to GC-TSG under a reasonable adaptation protocol, as described in~\ref{appendix:baseline}. For TimeGAN and TSGDiff, which do not natively support class-conditional generation, we train separate models for individual classes.

\subsection{Baseline Implementation for \gctsg}
\label{appendix:baseline}

Directly realizing Eq.~\eqref{eq:Chain_Rule} in the raw time-series space is fundamentally challenging. Each refinement step -- \ie $p_{\theta}(\mathbf{x}^{(\ell)}|\mathbf{x}^{(<\ell)})$ -- requires the model to condition on the full trajectory of all preceding coarser time-series representations, which makes training computationally expensive and often unstable for long time series. Common workarounds -- such as modeling $p_{\theta}(\mathbf{x}^{(j)}|\mathbf{x}^{(i)})$ for all granularity pairs -- fail to impose an explicit hierarchical ordering over temporal granularities, which may struggle to enforce consistent coarse-to-fine refinement or to capture dependencies across distant temporal granularities. 

To empirically substantiate these considerations, we present a comparative study of
baseline implementations that realize \gctsg in different representational spaces. Among them, the \textbf{AR Transformer} directly realizes Eq.~\eqref{eq:Chain_Rule} in the raw time-series space by modeling the full coarse-to-fine refinement trajectory, whereas the remaining baselines
approximate \gctsg by modeling the pairwise conditional $p_{\theta}(\mathbf{x}^{(j)} \mid \mathbf{x}^{(i)})$ across granularity pairs. Specifically, we examine three variants that differ in where the hierarchical refinement is
instantiated. A detailed comparison with the AR Transformer is provided in
Table~\ref{tab:architecture_ablation_conditional} for conditional generation and
Table~\ref{tab:architecture_ablation_marginal} for unconditional (marginal) generation.

\begin{itemize}

\item \textbf{\gctsg in raw time-series space.}

\define{RNN-TF/PF}
We implement RNN-TF and RNN-PF baselines using an encoder-decoder architecture with Gated Recurrent Units (GRUs) and a Mixture Density Network (MDN)~\citep{bishop_MDN_1994} head. The models map the target granularity level $\ell \in \{1, \dots, 8\}$ to a learnable embedding vector $\mathbf{e}_\ell$, which is injected into the decoder as a global condition. To incorporate the coarse condition $\mathbf{c}$, the encoder summarizes the sequence to initialize the decoder's hidden state. Additionally, at each time step $t$, the decoder receives a concatenated input vector consisting of the previous output $x_{t-1}$, the level embedding $\mathbf{e}_\ell$, and the local condition value $c_t$. The MDN head maps the decoder's hidden states to the parameters of a Gaussian Mixture Model for probabilistic generation. \textbf{RNN-TF} is trained via teacher forcing to minimize the negative log-likelihood, while \textbf{RNN-PF} employs adversarial training with a Bi-directional GRU discriminator to align hidden state dynamics and mitigate exposure bias. For marginal generation, $\mathbf{c}$ is replaced with a zero-valued sequence.

\define{Diffusion-TS} We extend the Diffusion-TS by incorporating the target granularity level and a coarse-condition. The model encodes the target granularity level $\ell \in \{1,\ldots,8\}$ into a vector embedding $\mathbf{e}_\ell$ using an MLP, which is injected into the Transformer backbone via Adaptive Layer Normalization alongside timestep embeddings. To incorporate the coarse condition $\mathbf{c}$, we concatenate it with the noisy input along the feature dimension. The model is trained to reconstruct the target time series at level $\ell$ from the noisy input, conditioned on $\mathbf{e}_\ell$ and $\mathbf{c}$. For marginal generation, conditioned on $\mathbf{e}_\ell$ alone, we replace $\mathbf{c}$ with a learnable \texttt{Null} embedding and train to reconstruct the target time series at level $\ell$.

\define{AR Transformer} We train an autoregressive Transformer with the MDN head on the concatenated 8-level sequence $\mathbf{s} = \mathbf{x}^{(1)} \oplus \ldots \oplus \mathbf{x}^{(8)} \in \mathbb{R}^{8T}$. The model uses learnable positional embeddings. To indicate level transitions, we insert a learnable [SEP] token at each level boundary with corresponding level embedding $\mathbf{e}_\ell$ added to [SEP] (and [BOS] for level 1). For unconditional generation, we begin from [BOS] and generate sequentially up to the desired target level. For conditional generation, given a condition at level $i$, the model requires the full trajectory up to that level $i$ (i.e., $\mathbf{x}^{(1)}, \ldots, \mathbf{x}^{(i)}$) as context. Unlike \name, where coarser levels are naturally obtained during tokenization, the AR Transformer cannot derive them from a single-level condition. Therefore, we provide the ground-truth trajectory $\{\mathbf{x}^{(\ell)}\}_{\ell=1}^{i}$ as a prefix and generate levels $i{+}1$ through the target level autoregressively. 

\item \textbf{\gctsg in continuous latent space.}

\define{TimeGAN} We extend the TimeGAN framework to support granularity-controllable generation. The generator receives $\boldsymbol{\epsilon}\oplus \mathbf{c} \oplus \mathbf{e}_\ell$ as input, where $\boldsymbol{\epsilon}$ is random noise, $\mathbf{c}$ is the coarse condition sequence, and $\mathbf{e}_\ell$ is the target level embedding. Unlike standard GANs, the generator produces a sequence of embeddings in the latent space, which is then mapped to the time-series space via a recovery network. The discriminator also operates in the latent space, distinguishing real latent representations $\mathbf{h}$ (derived from real time series via an embedding network) from generated ones $\hat{\mathbf{h}}$ using the concatenated input $\mathbf{h} \oplus \mathbf{c} \oplus \mathbf{e}_\ell$. To enable marginal generation, we omit the coarse condition (\ie $\mathbf{c} = \mathbf{0}$) for unconditional training samples, allowing the model to generate samples conditioned solely on the target granularity level.

\item \textbf{\gctsg in discrete token space.}

\define{TimeVQVAE} We adapt the two-stage TimeVQVAE framework for granularity-controllable generation. In Stage 1, we train the VQ-VAE tokenizer on time series at all granularity levels $\ell \in \{1, \ldots, 8\}$ by randomly sampling $\mathbf{x}^{(\ell)}$ during training, enabling the tokenizer to learn a unified discrete representation across varying granularities. In Stage 2, we extend the MaskGIT~\citep{chang2022maskgit} by concatenating condition tokens, a [SEP] token, and masked target tokens. To encode the target granularity level, we add a learnable level embedding $\mathbf{e}_\ell$ to the [SEP] token. For marginal generation, we replace the condition tokens with a learnable \texttt{Null} embedding and generate conditioned on [SEP]$+\mathbf{e}_\ell$ alone. For conditional generation, we tokenize the condition time series and generate conditioned on both the condition prefix and [SEP]$+\mathbf{e}_\ell$.

\define{TOTEM}
We adapt the two-stage TOTEM framework for granularity-controllable generation. In Stage 1, we train a VQ-VAE tokenizer that directly processes raw time series across all granularity levels $\ell \in \{1, \dots, 8\}$, enabling the model to learn a unified discrete representation for both high-resolution targets and coarsified signals. In Stage 2, we implement an Autoregressive Transformer to handle the generative process. The transformer receives a prefix consisting of the tokens from the source condition time series and a [SEP] token, where the target level information is integrated into the [SEP] token via a learnable level embedding $\mathbf{e}_\ell$. During training, the model is optimized using Teacher Forcing to predict the subsequent target tokens sequentially in an autoregressive manner. For marginal generation, the source condition sequence is replaced by a zero-valued input, and the model generates the target time series based on the [SEP] token and its associated level information alone.

\end{itemize}

\section{Limitations and Broader Impacts}
\label{app:limitations_broaderImpacts}
\define{Limitations} This work introduces \gctsg as a flexible generation task that extends conventional time-series generative modeling by explicitly controlling both input and output temporal granularity. To establish a clear and controlled formulation, our main experiments focus primarily on the single-dataset setting, where granularity levels can be defined through a clear hierarchy of information-reducing operators. Although we further provide an extension of \name to heterogeneous multi-domain data in~\cref{subsec:foundation} and~\cref{appendix:Extension_FM}, our evaluation of foundation-scale tokenization remains preliminary. A more comprehensive study involving broader downstream tasks, additional domains, and practical use cases is an important direction for future work.

Our experiments target univariate time series, and multivariate extensions remain future work. Each model is trained at a fixed sequence length, so results across different lengths do not establish mixed-length inference within a single model. Furthermore, our cross-domain evaluation concerns tokenizer transferability rather than zero-shot refinement by the full generative model, which remains unverified.
Refinements should be interpreted as plausible samples conditioned on coarse observations, rather than recovery of the unknown ground-truth fine-scale trajectory.

\define{Broader Impacts} \name has the potential to support a wide range of applications involving time-series data, particularly in domains where practitioners need to access, analyze, or generate data across multiple temporal granularities. Nevertheless, generated samples should undergo expert validation before being used for decision-making or downstream deployment, especially in high-stakes domains such as healthcare and finance.

\newpage

\section{Related Works}
 \label{app:related_work}
\subsection{Time-Series Generation}
\label{app:related_TSG}

Deep generative models have achieved strong performance across modalities, but time-series generation remains challenging because temporal data exhibit long-range dependencies, local fluctuations, nonstationarity, and domain-specific dynamics. TimeGAN~\citep{yoon_timegan_2019NeurIPS} adapts generative adversarial networks to time series by using supervised forecasting signals to improve temporal coherence. TimeVQVAE~\citep{Daesoo_TimeVQVAE_2023AISTATS} learns discrete time-series representations through vector quantization and performs generation in the resulting latent space. Diffusion-based methods have further improved sample quality through specialized inductive biases, including frequency-domain losses in Diffusion-TS~\citep{yuandiffusion}, graph-based temporal modeling in TSGDiff~\citep{shen_tsgdiff_AAAI2026}, and image-based representations in ImagenTime~\cite{naiman_ImagenTime_NeurIPS2024}.

These methods address the standard generation problem: learning to sample time series at the fixed temporal granularity specified by the training data. They do not natively expose temporal granularity as a controllable variable. In particular, they do not learn a single conditional family $p(x^{(j)}\mid x^{(i)})$ over arbitrary input-output granularity pairs, nor do they provide a mechanism for refining a coarse observation or user sketch into plausible finer-scale trajectories. Extending them to this setting requires adding external conditioning mechanisms, training separate pairwise models, or changing the architecture. TimeTok instead treats granularity control as part of the generative objective itself.

This distinction is important because GC-TSG is not only standard generation with a different sampling rate. It requires a model that can preserve coarse structure while introducing statistically plausible finer dynamics, and that can generate at different target granularities from the same learned representation. Thus, the central question is not only whether generated samples look realistic at one resolution, but whether a model can traverse a learned coarse-to-fine temporal hierarchy while maintaining consistency with the conditioning signal.

\subsection{Multi-Granular and Discrete Time-Series Representation Learning} 
\label{app:related_MG}
Recent time-series tokenization methods have shown that discrete representations can support scalable and general-purpose temporal modeling. TOTEM~\cite{talukdertotem} learns tokenized embeddings for time-series analysis, while WaveToken~\cite{masserano_wavetok_2025ICML} uses wavelet decompositions to capture time-localized frequency structure. TransPL~\cite{kimtranspl} decomposes temporal patches into trend and residual components and quantizes them using separate codebooks. These methods introduce useful structure into the latent space, but their tokenization is not designed as a controllable generative interface. The number of tokens is typically fixed or determined by the representation scheme, rather than by the desired output granularity. Consequently, they do not make prefix length semantically correspond to temporal granularity, and they do not couple tokenization with a generative process that can move from coarse to fine levels. 

Tokenization is also central to recent time-series foundation models. Chronos~\cite{ansari2024chronos} quantizes scaled observations into a fixed vocabulary using uniform binning, whereas Chronos-2~\cite{ansari2025chronos} embeds patches of continuous values together with time indices and observation masks. Neither scheme explicitly aligns token prefixes with temporal granularity levels. In contrast,  \name organizes tokens so that successive tokens carry progressively finer temporal granularity, making the token order itself a controllable interface over granularity.

\subsection{Multi-Scale Token Generation in Computer Vision} 
\label{app:related_MS}

Coarse-to-fine generation has been studied extensively in computer vision. VAR~\cite{tian_VAR_2024NeurIPS} and FlowAR~\cite{renflowar} replace token-by-token generation with next-scale generation: coarse visual tokens are generated first and condition subsequent finer-scale tokens. This has proven effective for improving image generation quality and scaling autoregressive image models. FlexTok~\cite{bachmann_flextok_2025ICML} takes a related but distinct direction by using nested dropout to make shorter token prefixes capture simpler image content, enabling efficient generation with variable token budgets. TimeTok is inspired by this broader idea of ordered token generation, but the problem it solves is different. In image generation, scale usually refers to spatial resolution, visual detail, or token efficiency. The generated object is still typically evaluated as a final full-resolution image, and intermediate scales are mainly a latent scaffold for better generation. In GC-TSG, temporal granularity is itself the user-facing object of control: the input may be a coarse time series, the output may be requested at any finer level, and intermediate levels correspond to different temporal dynamics rather than merely different visual resolutions.

This distinction is not cosmetic. Temporal granularity changes what aspects of the dynamics are observable: coarse levels preserve trends and long-range behavior, while fine levels recover transients, fluctuations, and local temporal events. Therefore, a time-series model must learn not only a nested representation, but a conditional refinement process over dynamics. Unlike VAR and FlowAR, TimeTok does not use the hierarchy only to improve final-resolution quality; it exposes the hierarchy as an input-output control mechanism. Unlike FlexTok, TimeTok explicitly models transitions between granularity levels through block-wise autoregressive generation, enabling conditional refinement from arbitrary coarse prefixes.

\subsection{Difference with Temporal Super-Resolution}
\label{app:related_TSR}
Temporal super-resolution reconstructs a high-resolution signal from low-resolution
measurements, and prior work typically treats it as a separate problem from 
time-series generation~\cite{duan2026srt}. TimeTok differs in two respects. First, the target granularity is 
a user-specified control variable, so a single model refines from any source level to any finer 
level rather than learning a fixed (low $\rightarrow$ high) mapping. Second, TimeTok also supports 
unconditional generation at any granularity, of which standard time-series generation is a special 
case. Super-resolution therefore corresponds to a single conditional instance of the broader \gctsg 
problem.

\subsection{Relation to Equivariance Learning}
\label{app:equivariance}

Equivariance learning encourages corresponding transformations in the latent and data spaces, for example by training downsampled latents to reconstruct correspondingly downsampled images~\cite{skorokhodov_EquivarianceLearning_ICML2025}. Our hierarchical decoding objective similarly
aligns token prefixes with corresponding temporal granularities. The key distinction is that TimeTok additionally learns generative transitions across granularities. While the aligned latent-to-data mapping does not itself generate finer latents from coarser ones, TimeTok's block-wise autoregressive model generates missing finer-level tokens conditioned on a coarse prefix. This enables generation at a selected granularity and refinement of coarse observations. The alignment therefore serves as a basis for
granularity-controllable generation, rather than the end goal.

\clearpage

\begin{table*}[p]
\centering

\begingroup
\scriptsize
\renewcommand{\arraystretch}{1.16}
\setlength{\tabcolsep}{2.8pt}

\rotatebox[origin=c]{270}{%
\begin{minipage}{0.96\textheight}
\centering
\caption{Comparison of \name with related lines of work.}
\label{tab:related_work_comparison_rotated}

\vspace{0.3em}

\begin{adjustbox}{max width=0.96\textheight, max totalheight=0.84\textwidth, keepaspectratio}
\begin{tabular}{
>{\raggedright\arraybackslash}p{2.75cm}
>{\raggedright\arraybackslash}p{3.05cm}
>{\centering\arraybackslash}p{1.75cm}
>{\centering\arraybackslash}p{2.05cm}
>{\centering\arraybackslash}p{2.05cm}
>{\centering\arraybackslash}p{2.05cm}
>{\centering\arraybackslash}p{2.25cm}
>{\raggedright\arraybackslash}p{5.65cm}
}
\toprule
\textbf{Line of work} &
\textbf{Representative methods} &
\textbf{Generates realistic time series} &
\textbf{Controls target granularity at inference} &
\textbf{Conditions on arbitrary coarser input} &
\textbf{Ordered prefix has temporal meaning} &
\textbf{Learns coarse-to-fine refinement trajectory} &
\textbf{Key distinction from \name{}} \\
\midrule

Standard time-series generation &
TimeGAN~\citep{yoon_timegan_2019NeurIPS}, Diffusion-TS~\citep{yuandiffusion}, TSGDiff~\citep{shen_tsgdiff_AAAI2026}, ImagenTime~\citep{naiman_ImagenTime_NeurIPS2024} &
Yes &
No &
No &
No &
No &
Learns samples at the training granularity; does not model $p(x^{(j)} \mid x^{(i)})$ across granularity pairs. \\
\midrule

Discrete time-series generation &
TimeVQVAE~\citep{Daesoo_TimeVQVAE_2023AISTATS} &
Yes &
No &
No &
No &
No &
Uses discrete latents, but token order does not define a controllable temporal hierarchy. \\
\midrule

Time-series tokenization / foundation representations &
TOTEM~\citep{talukdertotem} &
Representation, not primarily generation &
No &
No &
No &
No &
Learns useful tokens for downstream tasks, but not a generative interface for granularity control. \\
\midrule

Multi-granular time-series tokens &
WaveToken~\citep{masserano_wavetok_2025ICML}, TransPL~\citep{kimtranspl} &
Not the main goal &
No &
No &
Partly &
No &
Encodes multi-resolution or trend--residual structure, but granularity is fixed by the representation rather than chosen by the user. \\
\midrule

Multi-scale image generation &
VAR~\citep{tian_VAR_2024NeurIPS}, FlowAR~\citep{renflowar} &
Yes, for images &
Intermediate scales, mainly latent &
No &
Spatial/visual scale &
Yes, for image latent scales &
Uses coarse-to-fine token generation to improve final image quality; \name{} uses hierarchy as user-facing input-output control. \\
\midrule

Flexible image tokenization &
FlexTok~\citep{bachmann_flextok_2025ICML} &
Yes, for images &
Token budget / image complexity &
No &
Visual complexity &
No explicit temporal dynamics &
Variable prefixes control image complexity, not temporal refinement from coarse observations. \\
\midrule

\rowcolor[HTML]{F8FCFC}
Raw-space hierarchical GC-TSG baseline &
AR Transformer over $(x^{(1)}, \ldots, x^{(L)})$ &
Yes &
Yes &
Yes &
N/A &
Yes, but in raw space &
Directly models the right objective but is inefficient and weaker empirically; motivates \name{}'s token-space hierarchy. \\
\midrule

\rowcolor[HTML]{F0F8F8}
\textbf{\name{}} &
\textbf{This work} &
\textbf{Yes} &
\textbf{Yes} &
\textbf{Yes} &
\textbf{Yes} &
\textbf{Yes} &
\textbf{A single model for generation from scratch, generation at intermediate granularities, and conditional refinement from coarse inputs.} \\

\bottomrule
\end{tabular}
\end{adjustbox}
\end{minipage}%
}

\endgroup
\end{table*}
\clearpage
 \newpage

\section{ADDITIONAL EXPERIMENTAL RESULTS}
\label{app:Fullresult}
In this section, we present additional experiments that we omitted in the main body of the paper due to limited space.
\subsection{Compute Resource and Experiment Environment}
\label{app:compute_resource}
All experiments were conducted on NVIDIA A6000 GPU (48GB) or RTX PRO 6000 Blackwell~(96GB). We have used Python 3.10 to run all experiments, and will release the installation requirements through GitHub.

\subsection{Full Results}\label{app:full_results_table}

\begin{table*}[!ht]
\centering
\footnotesize
\setlength{\tabcolsep}{3.5pt}
\caption{\define{Marginal (Unconditional) Generation Performance of ECG5000} FID is evaluated at each granularity level (L1--L8).}
\label{tab:marginal_ecg5000}
\resizebox{\textwidth}{!}{%
\begin{tabular}{lcccccccc}
\toprule
\textbf{Method} 
& \textbf{L1} \color{blue}$\downarrow$
& \textbf{L2} \color{blue}$\downarrow$
& \textbf{L3} \color{blue}$\downarrow$
& \textbf{L4} \color{blue}$\downarrow$
& \textbf{L5} \color{blue}$\downarrow$
& \textbf{L6} \color{blue}$\downarrow$
& \textbf{L7} \color{blue}$\downarrow$
& \textbf{L8} \color{blue}$\downarrow$ \\
\midrule
RNN(TF)   & 2.778\std{1.096} & 1.393\std{0.500} & 0.459\std{0.129} & 1.121\std{0.430} & 0.643\std{0.150} & 0.206\std{0.064} & 0.286\std{0.101} & 0.188\std{0.057} \\
RNN(PF) & 0.649\std{0.106} & 0.485\std{0.079} & 0.202\std{0.036} & 0.170\std{0.042} & 0.118\std{0.044} & 0.105\std{0.026} & 0.199\std{0.023} & 0.082\std{0.021} \\
TimeGAN       & 0.139\std{0.024} & 0.224\std{0.036} & 0.263\std{0.093} & 0.477\std{0.143} & 0.989\std{0.186} & 0.872\std{0.358} & 1.474\std{0.414} & 1.870\std{0.296} \\
TimeVQVAE   & 0.045\std{0.011} & 0.031\std{0.005} & 0.017\std{0.002} & 0.017\std{0.004} & 0.031\std{0.003} & 0.021\std{0.004} & 0.036\std{0.009} & 0.034\std{0.011} \\
DiffusionTS   & 0.017\std{0.004} & 0.018\std{0.002} & 0.022\std{0.006} & 0.019\std{0.006} & 0.025\std{0.006} & 0.022\std{0.008} & 0.018\std{0.003} & 0.023\std{0.004} \\
TOTEM   & 0.025\std{0.005} & 0.020\std{0.001} & 0.024\std{0.005} & 0.015\std{0.004} & 0.028\std{0.010} & 0.022\std{0.005} & 0.039\std{0.009} & 0.058\std{0.016} \\
\midrule
\rowcolor[HTML]{F0F8F8} \textbf{\name}& \textbf{0.001\std{0.000}}& \textbf{0.001\std{0.001}}& \textbf{0.001\std{0.000}}& \textbf{0.001\std{0.000}}& \textbf{0.002\std{0.001}}& \textbf{0.003\std{0.001}}& \textbf{0.003\std{0.001}}& \textbf{0.006\std{0.001}}\\
\bottomrule
\end{tabular}
}%
\end{table*}

\begin{table*}[!ht]
\centering
\footnotesize
\setlength{\tabcolsep}{3.5pt}
\caption{\define{Marginal (Unconditional) Generation Performance of Nasdaq} FID is evaluated at each granularity level (L1--L8).}
\label{tab:marginal_fid_nasdaq}
\resizebox{\textwidth}{!}{%
\begin{tabular}{lcccccccc}
\toprule
\textbf{Method} 
& \textbf{L1} \color{blue}$\downarrow$
& \textbf{L2} \color{blue}$\downarrow$
& \textbf{L3} \color{blue}$\downarrow$
& \textbf{L4} \color{blue}$\downarrow$
& \textbf{L5} \color{blue}$\downarrow$
& \textbf{L6} \color{blue}$\downarrow$
& \textbf{L7} \color{blue}$\downarrow$
& \textbf{L8} \color{blue}$\downarrow$ \\
\midrule
RNN(TF)  & 8.029\std{0.629} & 7.490\std{0.606} & 8.406\std{1.007} & 8.173\std{0.654} & 8.859\std{0.154} & 11.34\std{1.167} & 11.05\std{2.312} & 3.362\std{0.317} \\
RNN(PF)  & 4.522\std{0.380} & 4.047\std{0.594} & 4.496\std{0.243} & 3.755\std{0.292} & 4.009\std{0.195} & 4.209\std{0.440} & 4.132\std{0.278} & 4.421\std{0.384} \\
TimeGAN   & 9.043\std{0.344} & 9.956\std{0.738} & 9.436\std{0.563} & 9.528\std{0.982} & 9.386\std{0.419} & 9.115\std{0.546} & 10.247\std{0.891} & 9.833\std{0.781} \\
TimeVQVAE  & 1.935\std{0.157} & 0.596\std{0.035} & 1.009\std{0.095} & 1.169\std{0.037} & 0.853\std{0.056} & 0.686\std{0.040} & 1.420\std{0.117} & 0.645\std{0.035} \\
DiffusionTS   & 0.198\std{0.019} & 0.252\std{0.019} & 0.322\std{0.040} & 0.369\std{0.028} & 0.539\std{0.035} & 0.440\std{0.077} & 0.597\std{0.103} & 0.739\std{0.063} \\
TOTEM  & 0.748\std{0.038} & 0.631\std{0.058} & 0.589\std{0.031} & 0.722\std{0.057} & 0.671\std{0.057} & 0.795\std{0.045} & 0.580\std{0.022} & 1.058\std{0.085} \\
\midrule
\rowcolor[HTML]{F0F8F8} 
\textbf{\name}& \textbf{0.088\std{0.010}} & \textbf{0.100\std{0.010}} & \textbf{0.121\std{0.027}} & \textbf{0.090\std{0.016}} & \textbf{0.106\std{0.013}} & \textbf{0.130\std{0.014}} & \textbf{0.094\std{0.007}} & \textbf{0.091\std{0.017}} \\
\bottomrule
\end{tabular}}%
\end{table*}

\begin{table*}[t]
\centering
\scriptsize
\caption{\define{Full Results of \gctsg} Full results for granularity-controlled time-series generation~(\gctsg) on the ECG5000, ItalyPowerDemand, Nasdaq, and ETTh1 datasets. Here, $i$ denotes the source level and $j$ the target level, \ie generating level-$j$ time series from level-$i$ input. We report $\text{CRPS}_{\text{sum}}$ and coarse-consistency (C-Cons), where C-Cons measures the RMSE between the original coarse input and the generated output after re-coarsening with a Gaussian filter.
}
\label{appendix_tab:conditional_GC_TSG}

\begin{tabular}{l| cc| cc| cc| cc}
\toprule
& \multicolumn{2}{c}{\textbf{ECG5000}} 
& \multicolumn{2}{c}{\textbf{ItalyPowerDemand}}
& \multicolumn{2}{c}{\textbf{Nasdaq}}
& \multicolumn{2}{c}{\textbf{ETTh1}}\\
\cmidrule(lr){2-3}\cmidrule(lr){4-5}\cmidrule(lr){6-7}\cmidrule(lr){8-9}

\textbf{$i\rightarrow j$}
& \textbf{$\text{CRPS}_{\text{sum}}$} \color{blue}$\downarrow$ & \textbf{C-Cons} \color{blue}$\downarrow$
& \textbf{$\text{CRPS}_{\text{sum}}$} \color{blue}$\downarrow$ & \textbf{C-Cons} \color{blue}$\downarrow$
& \textbf{$\text{CRPS}_{\text{sum}}$} \color{blue}$\downarrow$ & \textbf{C-Cons} \color{blue}$\downarrow$
& \textbf{$\text{CRPS}_{\text{sum}}$} \color{blue}$\downarrow$ & \textbf{C-Cons} \color{blue}$\downarrow$ \\
\midrule

$1\rightarrow 2$ &
1.079\stdtiny{0.408} & 0.011\stdtiny{0.004} &
2.785\stdtiny{0.307} & 0.020\stdtiny{0.005} &
20.09\stdtiny{16.54} & 0.091\stdtiny{0.073} &
21.44\stdtiny{34.86} & 0.375\stdtiny{2.135} \\

$1\rightarrow 3$ &
1.037\stdtiny{0.364} & 0.009\stdtiny{0.004} &
2.813\stdtiny{0.256} & 0.019\stdtiny{0.005} &
15.18\stdtiny{10.77} & 0.076\stdtiny{0.061} &
17.38\stdtiny{33.31} & 0.304\stdtiny{1.821} \\

$1\rightarrow 4$ &
1.374\stdtiny{0.568} & 0.010\stdtiny{0.007} &
2.644\stdtiny{0.307} & 0.016\stdtiny{0.007} &
15.02\stdtiny{12.54} & 0.081\stdtiny{0.079} &
17.15\stdtiny{35.46} & 0.338\stdtiny{2.063} \\

$1\rightarrow 5$ &
1.718\stdtiny{0.526} & 0.010\stdtiny{0.006} &
2.431\stdtiny{0.254} & 0.013\stdtiny{0.007} &
14.50\stdtiny{11.52} & 0.083\stdtiny{0.078} &
18.18\stdtiny{33.65} & 0.313\stdtiny{1.892} \\

$1\rightarrow 6$ &
2.173\stdtiny{0.674} & 0.011\stdtiny{0.007} &
2.037\stdtiny{0.262} & 0.009\stdtiny{0.006} &
16.48\stdtiny{12.47} & 0.095\stdtiny{0.088} &
19.45\stdtiny{30.63} & 0.297\stdtiny{1.798} \\

$1\rightarrow 7$ &
2.730\stdtiny{0.848} & 0.011\stdtiny{0.006} &
1.602\stdtiny{0.399} & 0.004\stdtiny{0.004} &
17.11\stdtiny{12.47} & 0.096\stdtiny{0.088} &
20.30\stdtiny{23.63} & 0.191\stdtiny{1.063} \\

$1\rightarrow 8$ &
3.042\stdtiny{0.900} & 0.011\stdtiny{0.007} &
1.061\stdtiny{0.277} & 0.002\stdtiny{0.002} &
17.72\stdtiny{12.63} & 0.089\stdtiny{0.085} &
23.25\stdtiny{25.00} & 0.220\stdtiny{1.191} \\

\midrule

$2\rightarrow 3$ &
0.969\stdtiny{0.354} & 0.010\stdtiny{0.004} &
2.850\stdtiny{0.241} & 0.021\stdtiny{0.005} &
14.55\stdtiny{9.16} & 0.075\stdtiny{0.056} &
17.98\stdtiny{35.96} & 0.354\stdtiny{2.118} \\

$2\rightarrow 4$ &
1.109\stdtiny{0.369} & 0.010\stdtiny{0.004} &
2.670\stdtiny{0.296} & 0.017\stdtiny{0.006} &
13.73\stdtiny{9.15} & 0.081\stdtiny{0.072} &
18.25\stdtiny{39.89} & 0.420\stdtiny{2.486} \\

$2\rightarrow 5$ &
1.485\stdtiny{0.556} & 0.010\stdtiny{0.006} &
2.380\stdtiny{0.274} & 0.013\stdtiny{0.006} &
14.57\stdtiny{10.87} & 0.090\stdtiny{0.082} &
18.86\stdtiny{38.85} & 0.398\stdtiny{2.416} \\

$2\rightarrow 6$ &
1.897\stdtiny{0.711} & 0.011\stdtiny{0.006} &
2.031\stdtiny{0.273} & 0.010\stdtiny{0.006} &
15.29\stdtiny{11.02} & 0.093\stdtiny{0.084} &
19.61\stdtiny{33.72} & 0.345\stdtiny{1.992} \\

$2\rightarrow 7$ &
2.401\stdtiny{0.924} & 0.011\stdtiny{0.006} &
1.607\stdtiny{0.398} & 0.004\stdtiny{0.004} &
15.58\stdtiny{11.03} & 0.088\stdtiny{0.082} &
20.45\stdtiny{26.13} & 0.226\stdtiny{1.193} \\

$2\rightarrow 8$ &
2.693\stdtiny{1.001} & 0.012\stdtiny{0.007} &
1.004\stdtiny{0.344} & 0.002\stdtiny{0.004} &
17.73\stdtiny{11.26} & 0.093\stdtiny{0.080} &
21.36\stdtiny{22.91} & 0.176\stdtiny{1.060} \\

$3\rightarrow 4$ &
1.022\stdtiny{0.589} & 0.011\stdtiny{0.006} &
2.704\stdtiny{0.286} & 0.019\stdtiny{0.005} &
13.94\stdtiny{9.85} & 0.090\stdtiny{0.078} &
14.25\stdtiny{29.15} & 0.233\stdtiny{1.688} \\

$3\rightarrow 5$ &
1.283\stdtiny{0.639} & 0.011\stdtiny{0.006} &
2.435\stdtiny{0.232} & 0.016\stdtiny{0.008} &
13.56\stdtiny{9.00} & 0.087\stdtiny{0.075} &
14.87\stdtiny{29.45} & 0.224\stdtiny{1.638} \\

$3\rightarrow 6$ &
1.637\stdtiny{0.776} & 0.012\stdtiny{0.007} &
2.081\stdtiny{0.289} & 0.012\stdtiny{0.006} &
13.93\stdtiny{10.06} & 0.084\stdtiny{0.079} &
16.16\stdtiny{25.64} & 0.203\stdtiny{1.354} \\

$3\rightarrow 7$ &
2.026\stdtiny{0.766} & 0.012\stdtiny{0.007} &
1.536\stdtiny{0.368} & 0.005\stdtiny{0.005} &
16.46\stdtiny{11.22} & 0.101\stdtiny{0.086} &
19.81\stdtiny{26.20} & 0.218\stdtiny{1.156} \\

$3\rightarrow 8$ &
2.322\stdtiny{0.874} & 0.012\stdtiny{0.007} &
0.864\stdtiny{0.285} & 0.002\stdtiny{0.002} &
17.82\stdtiny{12.09} & 0.102\stdtiny{0.089} &
22.03\stdtiny{26.28} & 0.216\stdtiny{1.179} \\

\midrule

$4\rightarrow 5$ &
0.988\stdtiny{0.360} & 0.010\stdtiny{0.005} &
2.442\stdtiny{0.214} & 0.018\stdtiny{0.008} &
13.44\stdtiny{9.77} & 0.094\stdtiny{0.085} &
16.00\stdtiny{31.17} & 0.272\stdtiny{1.541} \\

$4\rightarrow 6$ &
1.265\stdtiny{0.468} & 0.011\stdtiny{0.006} &
2.056\stdtiny{0.267} & 0.013\stdtiny{0.006} &
14.38\stdtiny{10.80} & 0.096\stdtiny{0.087} &
16.81\stdtiny{28.87} & 0.248\stdtiny{1.362} \\

$4\rightarrow 7$ &
1.602\stdtiny{0.591} & 0.012\stdtiny{0.007} &
1.513\stdtiny{0.371} & 0.006\stdtiny{0.004} &
15.81\stdtiny{12.10} & 0.105\stdtiny{0.095} &
20.64\stdtiny{29.00} & 0.274\stdtiny{1.295} \\

$4\rightarrow 8$ &
1.858\stdtiny{0.638} & 0.011\stdtiny{0.006} &
0.810\stdtiny{0.319} & 0.002\stdtiny{0.003} &
17.67\stdtiny{14.28} & 0.105\stdtiny{0.101} &
23.75\stdtiny{30.20} & 0.291\stdtiny{1.363} \\

\midrule

$5\rightarrow 6$ &
0.906\stdtiny{0.331} & 0.009\stdtiny{0.005} &
2.026\stdtiny{0.279} & 0.014\stdtiny{0.006} &
14.78\stdtiny{12.08} & 0.106\stdtiny{0.098} &
18.59\stdtiny{34.74} & 0.352\stdtiny{1.636} \\

$5\rightarrow 7$ &
1.195\stdtiny{0.396} & 0.009\stdtiny{0.005} &
1.504\stdtiny{0.390} & 0.007\stdtiny{0.004} &
15.34\stdtiny{12.95} & 0.104\stdtiny{0.101} &
19.28\stdtiny{29.45} & 0.275\stdtiny{1.288} \\

$5\rightarrow 8$ &
1.526\stdtiny{0.528} & 0.010\stdtiny{0.006} &
0.750\stdtiny{0.210} & 0.002\stdtiny{0.002} &
18.06\stdtiny{15.97} & 0.114\stdtiny{0.112} &
20.56\stdtiny{26.87} & 0.235\stdtiny{1.231} \\

\midrule

$6\rightarrow 7$ &
0.914\stdtiny{0.390} & 0.007\stdtiny{0.005} &
1.541\stdtiny{0.407} & 0.008\stdtiny{0.004} &
16.34\stdtiny{16.50} & 0.121\stdtiny{0.123} &
15.22\stdtiny{25.32} & 0.229\stdtiny{1.130} \\

$6\rightarrow 8$ &
1.332\stdtiny{0.502} & 0.008\stdtiny{0.005} &
0.705\stdtiny{0.186} & 0.003\stdtiny{0.004} &
17.81\stdtiny{16.46} & 0.120\stdtiny{0.119} &
18.21\stdtiny{25.78} & 0.236\stdtiny{1.206} \\

\midrule

$7\rightarrow 8$ &
0.835\stdtiny{0.378} & 0.006\stdtiny{0.005} &
0.664\stdtiny{0.186} & 0.004\stdtiny{0.003} &
17.53\stdtiny{16.12} & 0.129\stdtiny{0.126} &
16.70\stdtiny{27.14} & 0.252\stdtiny{1.262} \\

\bottomrule
\end{tabular}
\end{table*}





\begin{table*}[!ht]
\centering
\small
\caption{\define{Baseline Comparison of \gctsg} We compare \name with TOTEM (strongest baseline) in \gctsg task on the ItalyPowerDemand and ETTh1 datasets. We report the average~(Avg) of all $i\rightarrow j$ pairs. 
}
\label{appendix_tab:conditional_GC_TSG_2}
\begin{tabular}{c| cc| cc}
\toprule
& \multicolumn{2}{c|}{\textbf{ItalyPowerDemand}}
& \multicolumn{2}{c}{\textbf{ETTh1}}\\
\cmidrule(lr){2-3}\cmidrule(lr){4-5}
\textbf{$Avg$}
& \textbf{$\text{CRPS}_{\text{sum}}$} \color{blue}$\downarrow$ & \textbf{C-Cons} \color{blue}$\downarrow$
& \textbf{$\text{CRPS}_{\text{sum}}$} \color{blue}$\downarrow$ & \textbf{C-Cons} \color{blue}$\downarrow$ \\
\midrule
TOTEM &
1.099\stdtiny{0.761} & 0.074\stdtiny{0.041} &
29.45\stdtiny{13.14} & 0.358\stdtiny{0.219} \\
\midrule
\rowcolor[HTML]{F0F8F8} \textbf{\name}&
\textbf{0.501\stdtiny{0.357}} & \textbf{0.027\stdtiny{0.019}} &
\textbf{12.22\stdtiny{6.245}} & \textbf{0.104\stdtiny{0.084}} \\
\bottomrule
\end{tabular}
\end{table*}

\begin{table*}[!ht]
\centering
\small
\caption{\define{Baseline Comparison of Marginal Generation} We compare \name with TOTEM in marginal time series generation task on the ItalyPowerDemand and ETTh1 datasets across 8 layers (L1--L8). We report the FID score.
}
\label{appendix_tab:marginal}
\begin{tabular}{ll| cccc}
\toprule
\textbf{Method} & \textbf{Dataset} 
& \textbf{L1} \color{blue}$\downarrow$ & \textbf{L2} \color{blue}$\downarrow$ 
& \textbf{L3} \color{blue}$\downarrow$ & \textbf{L4} \color{blue}$\downarrow$ \\
\midrule
TOTEM & ItalyPowerDemand 
& 0.061\stdtiny{0.007} & 0.050\stdtiny{0.007} & 0.078\stdtiny{0.011} & 0.084\stdtiny{0.019} \\
\rowcolor[HTML]{F0F8F8} \textbf{\name}& ItalyPowerDemand 
& \textbf{0.006\stdtiny{0.000}} & \textbf{0.009\stdtiny{0.000}} & \textbf{0.006\stdtiny{0.000}} & \textbf{0.006\stdtiny{0.000}} \\
\midrule
TOTEM & ETTh1 
& 0.300\stdtiny{0.027} & 0.246\stdtiny{0.013} & 0.537\stdtiny{0.051} & 0.723\stdtiny{0.049} \\
\rowcolor[HTML]{F0F8F8} \textbf{\name}& ETTh1 
& \textbf{0.029\stdtiny{0.002}} & \textbf{0.034\stdtiny{0.001}} & \textbf{0.066\stdtiny{0.003}} & \textbf{0.038\stdtiny{0.003}} \\
\midrule
\midrule
\textbf{Method} & \textbf{Dataset} 
& \textbf{L5} \color{blue}$\downarrow$ & \textbf{L6} \color{blue}$\downarrow$ 
& \textbf{L7} \color{blue}$\downarrow$ & \textbf{L8} \color{blue}$\downarrow$ \\
\midrule
TOTEM & ItalyPowerDemand 
& 0.062\stdtiny{0.008} & 0.076\stdtiny{0.010} & 0.120\stdtiny{0.016} & 0.140\stdtiny{0.033} \\
\rowcolor[HTML]{F0F8F8} \textbf{\name}& ItalyPowerDemand 
& \textbf{0.007\stdtiny{0.000}} & \textbf{0.011\stdtiny{0.002}} & \textbf{0.009\stdtiny{0.001}} & \textbf{0.006\stdtiny{0.000}} \\
\midrule
TOTEM & ETTh1 
& 0.547\stdtiny{0.020} & 0.519\stdtiny{0.030} & 0.720\stdtiny{0.024} & 0.568\stdtiny{0.038} \\
\rowcolor[HTML]{F0F8F8} \textbf{\name}& ETTh1 
& \textbf{0.087\stdtiny{0.008}} & \textbf{0.093\stdtiny{0.007}} & \textbf{0.080\stdtiny{0.008}} & \textbf{0.105\stdtiny{0.014}} \\
\bottomrule
\end{tabular}
\end{table*}

\begin{table*}[t]
\centering
\small
\caption{\define{Additional Foundation Tokenizer Evaluation.}
Evaluation of the UTSD-pretrained tokenizers on ItalyPowerDemand and
ETTh1 under the same protocol as Table~6.
}
\label{tab:foundation_additional}
\begin{tabular}{lcccc}
\toprule
& \multicolumn{2}{c}{\textbf{ItalyPowerDemand}}
& \multicolumn{2}{c}{\textbf{ETTh1}} \\
\cmidrule(lr){2-3}\cmidrule(lr){4-5}
\textbf{Method}
& FID $\downarrow$
& ACC $\uparrow$
& FID $\downarrow$
& MSE $\downarrow$ \\
\midrule
TOTEM
& 0.0466
& 0.9545\stdtiny{0.0164}
& 0.0916
& 0.1098\stdtiny{0.0002} \\
\midrule
\rowcolor[HTML]{F0F8F8}
\name (w/o DFA)
& 0.0784
& 0.9788\stdtiny{0.0026}
& 0.2925
& 0.1057\stdtiny{0.0005} \\
\rowcolor[HTML]{F0F8F8}
\textbf{\name (w/ DFA)}
& \textbf{0.0388}
& \textbf{0.9818\stdtiny{0.0045}}
& \textbf{0.0593}
& \textbf{0.1013\stdtiny{0.0004}} \\
\bottomrule
\end{tabular}
\end{table*}

\newpage
\subsection{Ablation Study}
\label{app:Ablation}

\begin{table*}[!ht]
\centering
\scriptsize
\setlength{\tabcolsep}{3.5pt}
\caption{\define{Conditional Generation}
Conditional GC-TSG evaluation under two settings:
Coarse-to-Fine ($i=1$, $j=8$) and All-Pairs average~(Avg, $i\rightarrow j$, $0<i< j$). C-Cons denotes coarse-level consistency, measured by RMSE.}
\label{tab:architecture_ablation_conditional}
\resizebox{0.97\textwidth}{!}{%
\begin{tabular}{c|cc |cc |cc |cc}
\toprule
& \multicolumn{4}{c}{\textbf{Classification (ECG5000)}} 
& \multicolumn{4}{c}{\textbf{Forecasting (Nasdaq)}}\\
\cmidrule(lr){2-5}\cmidrule(lr){6-9}

& \multicolumn{2}{c}{\textbf{C2F ($1\rightarrow 8$)}} 
& \multicolumn{2}{c}{\textbf{C2F-All Pairs ($i\rightarrow j$)}} 
& \multicolumn{2}{c}{\textbf{C2F ($1\rightarrow 8$)}} 
& \multicolumn{2}{c}{\textbf{C2F-All Pairs ($i\rightarrow j$)}}\\
\cmidrule(lr){2-3}\cmidrule(lr){4-5}\cmidrule(lr){6-7}\cmidrule(lr){8-9}

\textbf{Method}
& \textbf{$\text{CRPS}_{\text{sum}}$} \color{blue}$\downarrow$
& \textbf{C-Cons} \color{blue}$\downarrow$
& \textbf{$\text{CRPS}_{\text{sum}}$} \color{blue}$\downarrow$
& \textbf{C-Cons} \color{blue}$\downarrow$
& \textbf{$\text{CRPS}_{\text{sum}}$} \color{blue}$\downarrow$
& \textbf{C-Cons} \color{blue}$\downarrow$
& \textbf{$\text{CRPS}_{\text{sum}}$} \color{blue}$\downarrow$
& \textbf{C-Cons} \color{blue}$\downarrow$ \\
\toprule

AR Transformer&
4.827\stdtiny{1.768}& 0.027\stdtiny{0.014}&
5.804\stdtiny{1.658}& 0.040\stdtiny{0.018}& 32.57\stdtiny{25.82}&
0.160\stdtiny{0.136}& 22.98\stdtiny{18.95}& 0.119\stdtiny{0.107}\\

\midrule
\rowcolor[HTML]{F0F8F8} 
\textbf{\name}&
\textbf{3.042\stdtiny{0.900}} & \textbf{0.011\stdtiny{0.007}}&
\textbf{1.587\stdtiny{0.867}} & \textbf{0.010\stdtiny{0.006}} & \textbf{17.72\stdtiny{12.63}} &
\textbf{0.089\stdtiny{0.085}} & \textbf{15.87\stdtiny{12.48}}  & \textbf{0.096\stdtiny{0.090}} \\
\bottomrule
\end{tabular}
}
\end{table*}

\begin{table*}[!ht]
\centering
\footnotesize
\setlength{\tabcolsep}{3.5pt}
\caption{
Unconditional (marginal) generation performance across granularity levels (L1--L8), evaluated on ECG5000 and Nasdaq.}
\label{tab:architecture_ablation_marginal}
\resizebox{\textwidth}{!}{%
\begin{tabular}{llcccccccc}
\toprule
\textbf{Dataset} & \textbf{Method}
& \textbf{L1} \color{blue}$\downarrow$
& \textbf{L2} \color{blue}$\downarrow$
& \textbf{L3} \color{blue}$\downarrow$
& \textbf{L4} \color{blue}$\downarrow$
& \textbf{L5} \color{blue}$\downarrow$
& \textbf{L6} \color{blue}$\downarrow$
& \textbf{L7} \color{blue}$\downarrow$
& \textbf{L8} \color{blue}$\downarrow$ \\
\midrule
ECG5000
& AR Transformer
& 1.060\std{0.358}& 1.713\std{0.431}& 1.420\std{0.469}& 2.019\std{0.693}& 2.918\std{0.804}& 3.213\std{0.684}& 3.494\std{0.736}& 3.938\std{0.921}\\
\midrule
Nasdaq
& AR Transformer
& 14.990\std{0.730}& 12.427\std{0.886}& 13.305\std{1.111}& 13.287\std{0.868}& 12.870\std{0.852}& 13.309\std{0.622}& 13.443\std{1.722}& 12.405\std{1.176}\\
\bottomrule
\end{tabular}
}%
\end{table*}



\paragraph{\textbf{Token Allocation Across Granularity Levels}.}

\begin{figure}[!ht]
    \centering
    \includegraphics[width=\linewidth]{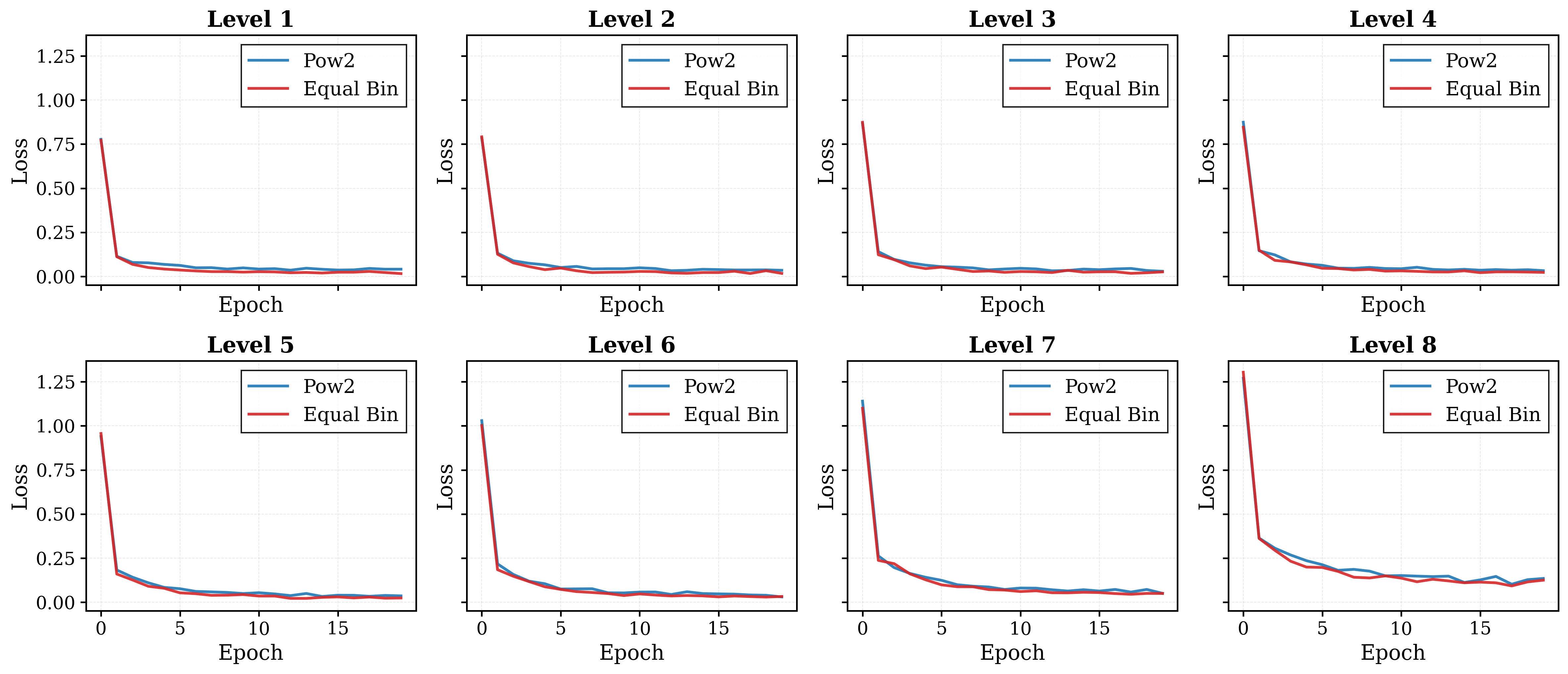}
    \caption{
       Level-wise training loss comparison between exponential (Pow2) and uniform (Equal Bin) token allocation strategies across all eight granularity levels. Both strategies exhibit nearly identical optimization behavior at every level, indicating that allocating fewer tokens to coarser levels does not adversely affect training dynamics. The x-axis denotes training progress, where one unit corresponds to 10 epochs.}
    \label{appendix:token_ablation}
\end{figure}

\begin{figure}[!ht]
    \centering
    \includegraphics[width=0.5\linewidth]{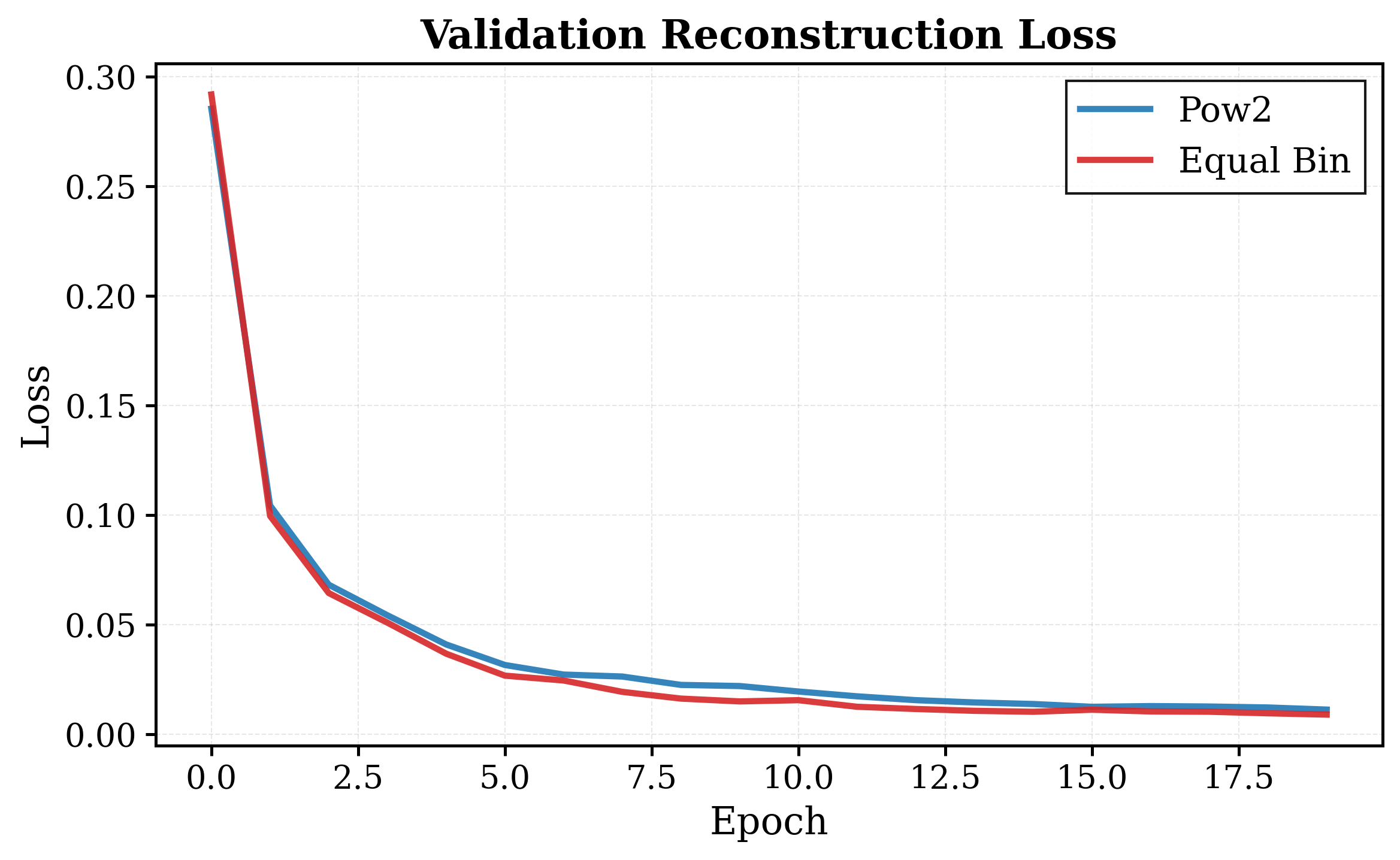}
    \caption{
       Validation reconstruction loss measured on the original time-series signals, comparing exponential (Pow2) and uniform (Equal Bin) token allocation strategies. The two approaches show nearly identical convergence trends, suggesting comparable reconstruction performance on the validation set.}
    \label{appendix:token_ablation_val}
\end{figure}

We study how the conditioning budget affects optimization by comparing our default exponential allocation $n_\ell = 2^{\ell-1}$ against a uniform allocation $n_\ell = 16 \cdot \ell$, while fixing the maximum granularity level to $L=8$ and the total finest-level budget to $n_L=M=128$. As shown in \cref{appendix:token_ablation}, the two token allocation strategies exhibit nearly identical training dynamics across all levels. Consistently, validation reconstruction losses on the original time-series data --~measured as the mean squared error between generated sequences and the ground truth~--follow similar trajectories under both allocations, as seen in \cref{appendix:token_ablation_val}. These results suggest that coarse time-series representations require substantially fewer tokens to achieve comparable optimization and reconstruction behavior, making exponential token allocation an efficient design choice under a fixed budget.

\paragraph{Comparison with AR Transformer in Raw Time-Series Space.}
We compare our approach with an AR Transformer that directly operates in the raw
time-series space and explicitly models the full coarse-to-fine refinement trajectory according to
Eq.~\eqref{eq:Chain_Rule}. Unlike our method, which approximates refinement via conditional
generation across granularity pairs in a structured token space, the AR Transformer must condition on the entire refinement history, leading to unstable learning. The results reveal a clear contrast between conditional and marginal generation behaviors of the AR Transformer. As shown in Table~\ref{tab:architecture_ablation_conditional}, the AR Transformer demonstrates reasonable performance in the conditional setting, where external conditions provide sufficient guidance to partially stabilize the generation process. However, as evidenced in Table~\ref{tab:architecture_ablation_marginal}, its performance degrades substantially in the
marginal (unconditional) generation setting. This failure highlights the intrinsic difficulty of
learning and sampling the entire coarse-to-fine refinement trajectory in the raw time-series
space. 

\subsection{Granularity Monotonicity and Smoothing Supervision}
\label{app:granularity_monotonicity}

We evaluate granularity ordering on 274 HandOutlines samples
(length 2,709) across eight token budgets ($1,2,4,\ldots,128$).
The mean Spearman correlation between increasing token budgets
and decreasing DFA exponents is 0.966.
The adjacent-level violation rate, where the DFA exponent increases
despite a larger token budget, is 13.5\%.
Of these violations, 86.9\% occur at the one-to-two-token transition;
violation rates for transitions from four through 128 tokens
are at most 0.7\%.
When every prefix instead reconstructs the original signal without
smoothing supervision, the mean correlation is near zero and
the violation rate increases to 47.1\%.
These results support the importance of smoothing supervision
for learning an ordered granularity hierarchy, although strict
monotonicity is not guaranteed.

\subsection{Long-Sequence Generation and Model Size}
\label{subsec:long_sequence}

\begin{table*}[t]
\centering
\begin{minipage}[t]{0.6\textwidth}
\vspace{0pt}
\centering
\caption{\define{Long-Sequence Generation}
HandOutlines results (length 2,709).
Oracle uses real training data.}
\label{tab:long_sequence}
\small
\setlength{\tabcolsep}{4pt}
\renewcommand{\arraystretch}{1.3}
\begin{tabularx}{\linewidth}{l
  >{\centering\arraybackslash}X
  >{\centering\arraybackslash}X
  >{\centering\arraybackslash}X}
\toprule
\textbf{Method}
& \textbf{FID} $\downarrow$
& \textbf{ACC} $\uparrow$
& \textbf{F1} $\uparrow$ \\
\midrule
Oracle
& --
& 0.920\stdtiny{0.019}
& 0.913\stdtiny{0.019} \\
\midrule
Diffusion-TS
& 0.621\stdtiny{0.143}
& 0.646\stdtiny{0.003}
& 0.420\stdtiny{0.015} \\
TimeVQVAE
& 0.041\stdtiny{0.0035}
& 0.675\stdtiny{0.005}
& 0.586\stdtiny{0.022} \\
\rowcolor[HTML]{F0F8F8}
\textbf{\name}
& \textbf{0.023\stdtiny{0.001}}
& \textbf{0.760\stdtiny{0.024}}
& \textbf{0.712\stdtiny{0.055}} \\
\bottomrule
\end{tabularx}
\end{minipage}%
\hfill
\begin{minipage}[t]{0.37\textwidth}
\vspace{0pt}
\centering
\caption{\define{Model Parameter Counts}
Parameters (M) on Nasdaq and HandOutlines.}
\label{tab:parameter_count}
\small
\setlength{\tabcolsep}{4pt}
\renewcommand{\arraystretch}{1.0}
\begin{tabular*}{\linewidth}{@{\extracolsep{\fill}}lcc@{}}
\toprule
\textbf{Method}
& \textbf{Nasdaq}
& \textbf{HandOutlines} \\
\midrule
TimeVQVAE   & 1.8  & 16.1 \\
TSGDiff     & 26.3 & 30.2 \\
Diffusion-TS & 0.6 & 51.1 \\
ImagenTime  & 57.6 & 151.7 \\
\rowcolor[HTML]{F0F8F8}
\textbf{\name} & 9.42 & \textbf{9.84} \\
\bottomrule
\end{tabular*}
\end{minipage}
\end{table*}

We evaluate \name on HandOutlines (length 2,709) to examine
generation beyond the maximum length of 256 used in our main
experiments.
As shown in Table~\ref{tab:long_sequence}, \name outperforms
Diffusion-TS and TimeVQVAE across FID, classification accuracy,
and F1, demonstrating its applicability to substantially longer
sequences.
Each model is trained at the evaluated sequence length;
this experiment does not establish mixed-length inference
within a single trained model.

\subsection{Model Parameter Count}
\label{app:parameter_count}
Table~\ref{tab:parameter_count} compares parameter counts on
Nasdaq (length 244) and HandOutlines (length 2,709).
\name maintains a similar model size of 9.42M--9.84M parameters.
Although TimeVQVAE and Diffusion-TS are smaller on Nasdaq,
\name uses fewer parameters than all compared methods
on HandOutlines while achieving strong generation performance.

\vfill
\clearpage
\newpage
\section{Coarse To Fine Reconstruction}
\label{appendix:full_coarse_2_fine}
We compare the coarse-to-fine \emph{reconstruction} performance between individual tokenizers~(I) trained on specific datasets and a foundation tokenizer~(F) trained on the UTSD dataset, which contains time series from various domains.
\begin{figure*}[!ht]
\begin{center}
\center{\includegraphics[width=\textwidth]{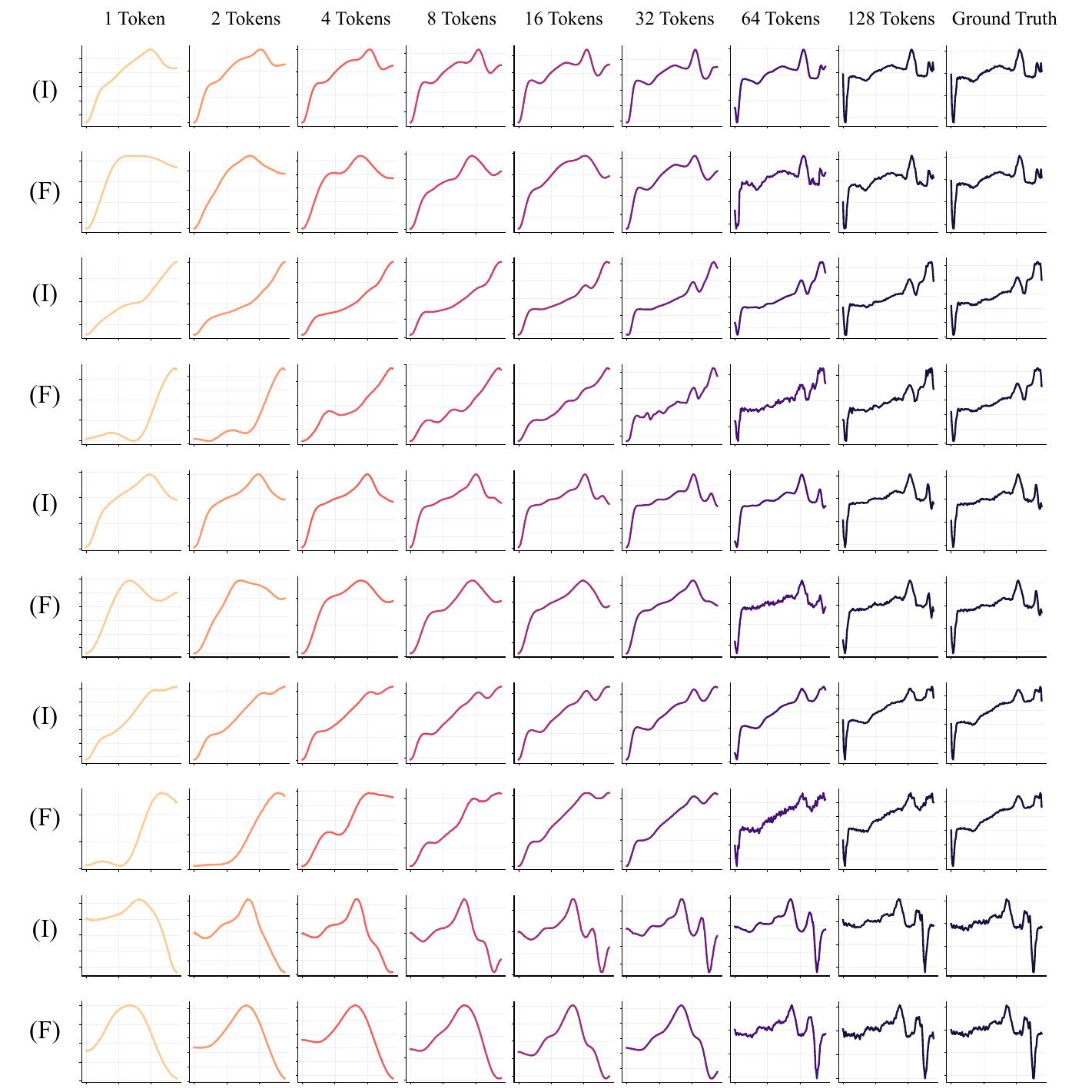}}
\caption{\define{ECG5000 Reconstruction}  }
\label{appendix_fig:ecg5000_recon}
\end{center}
\end{figure*}

\begin{figure*}[!ht]
\begin{center}
\center{\includegraphics[width=\textwidth]{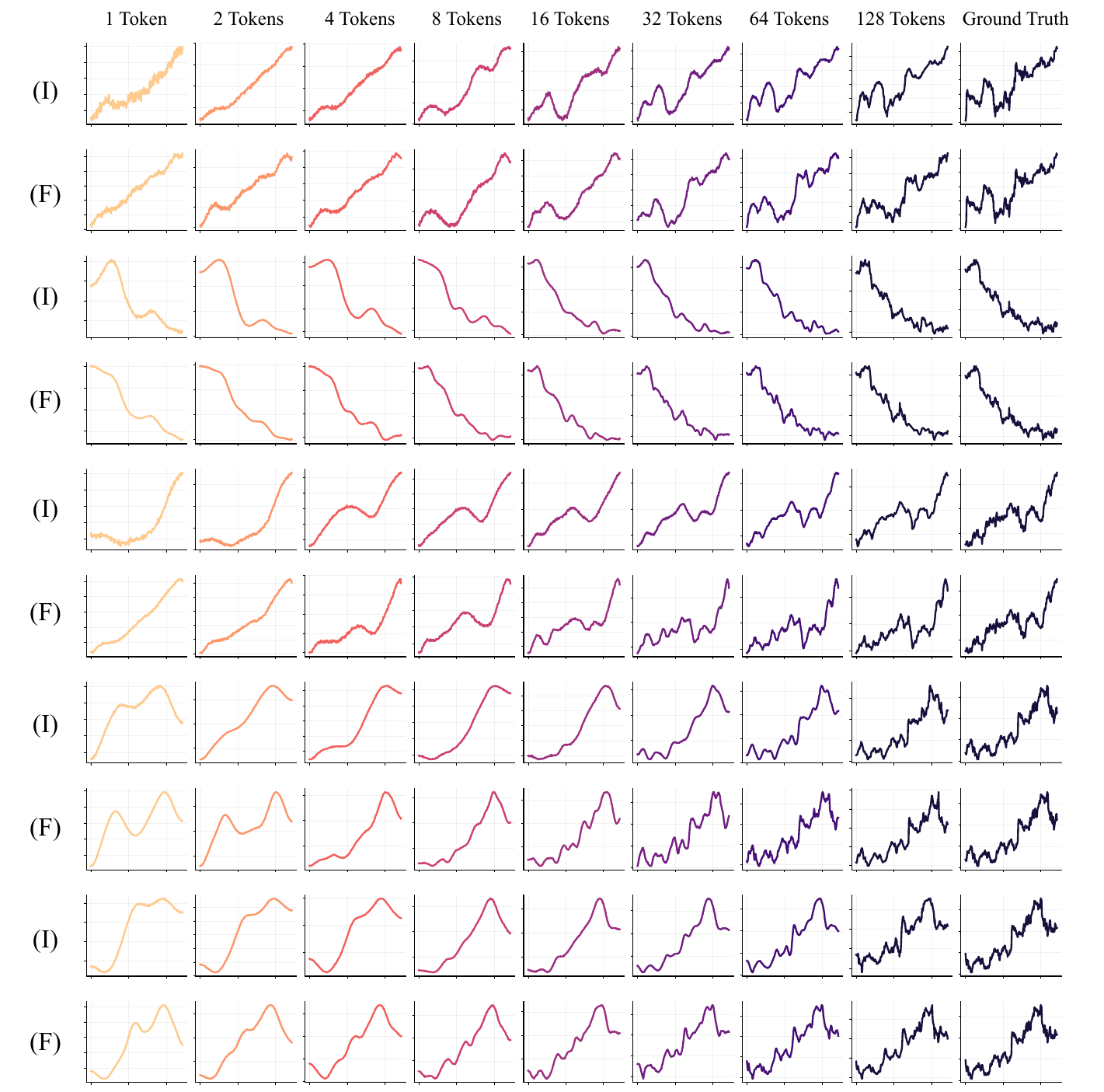}}
\caption{\define{Nasdaq Reconstruction}}
\label{appendix_fig:Nasdaq_recon}
\end{center}
\end{figure*}

\vfill
\clearpage
\newpage
\section{\gctsg Generation}
\label{appendix:gctsg_full}
\begin{figure*}[!ht]
\begin{center}
\center{\includegraphics[width=0.83\textwidth]{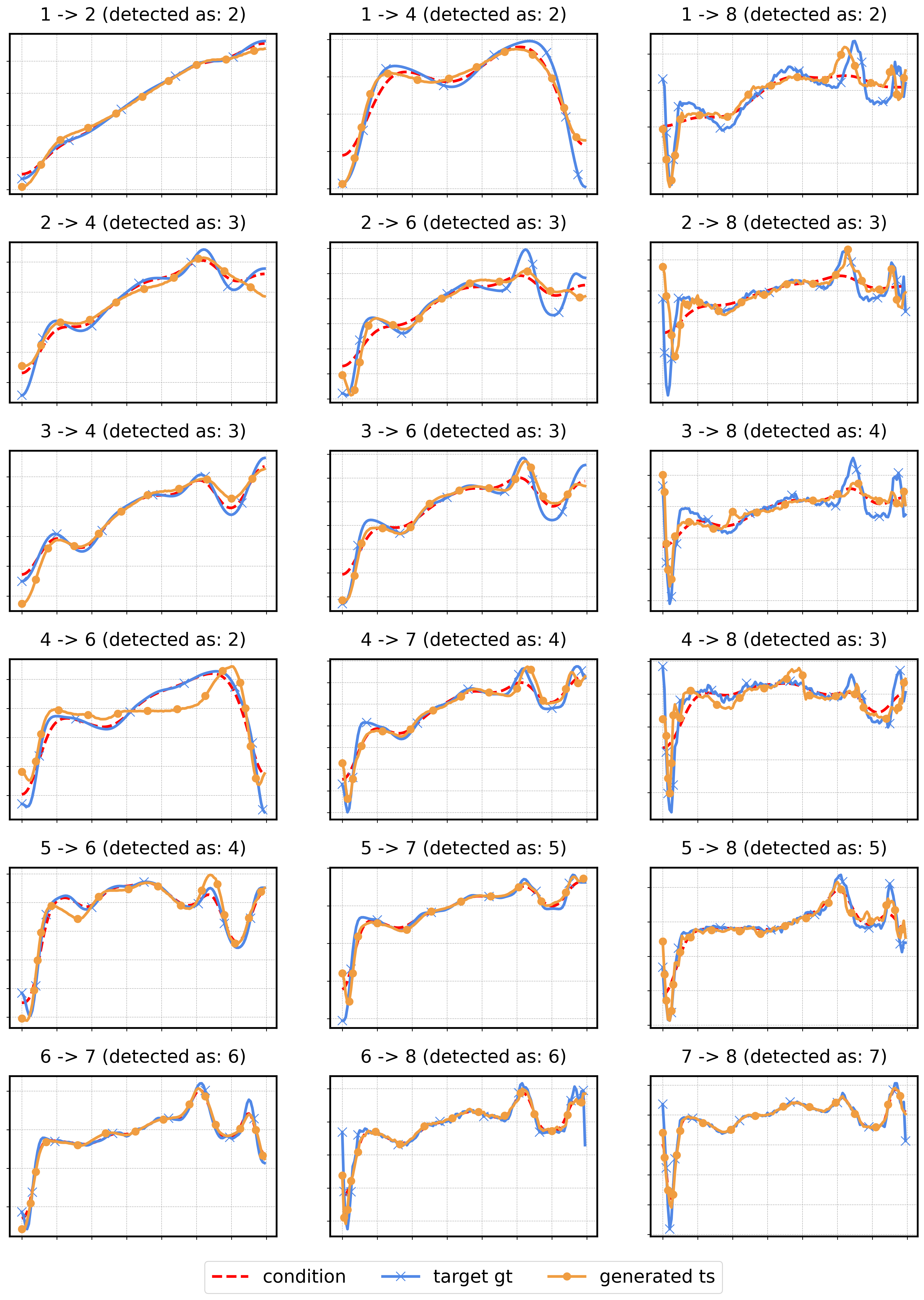}}
\caption{\define{\gctsg of ECG5000} We visualize the generation of $i\rightarrow j$ in ECG5000 dataset. Our level detection module first detects the granularity level of the sample. Using this sample as the condition, we generate time series at target granularity.}
\label{appendix_fig:gctsg_ecg5000}
\end{center}
\end{figure*}

\newpage 
\section{Other Information Reducing Operations}
\label{appendix:info_reducing}

\begin{figure*}[!ht]
\begin{center}
\center{\includegraphics[width=\textwidth]{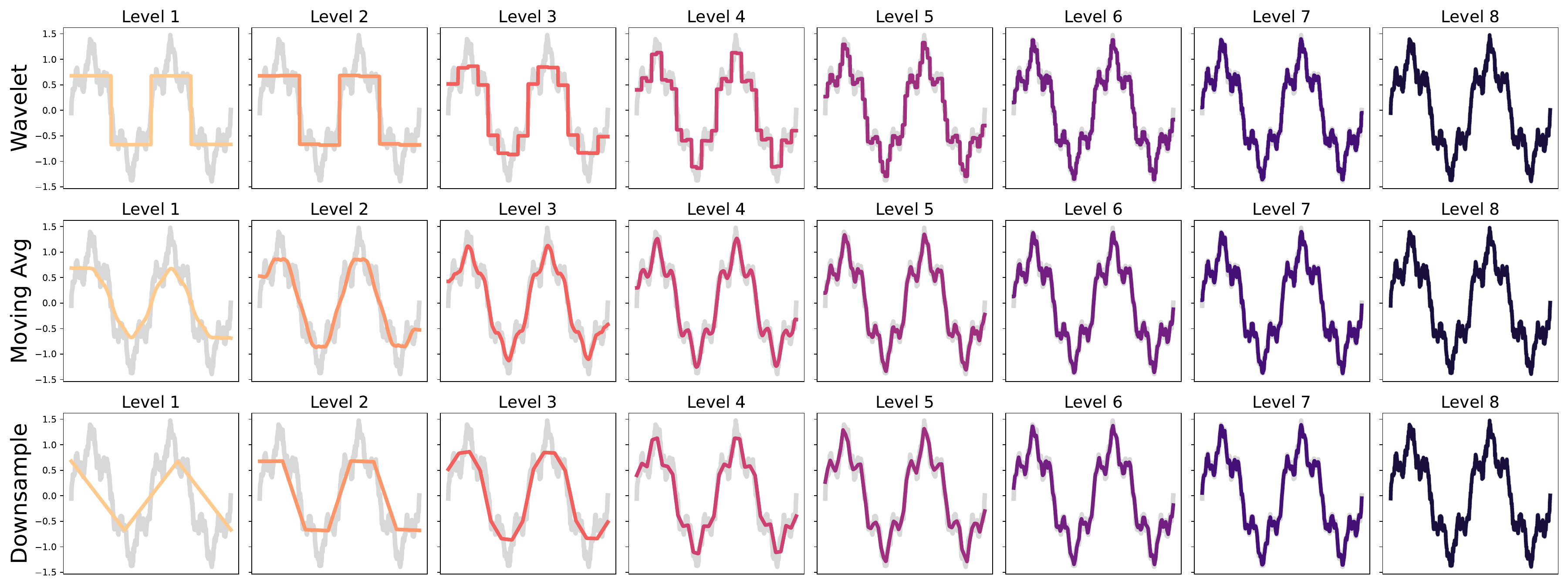}}
\caption{\define{Information Reducing Operators} Visualization of alternative information reducing operators for each granularity level}
\label{appendix_fig:alternatives}
\end{center}
\end{figure*}


\begin{table*}[!ht]
\centering
\scriptsize
\setlength{\tabcolsep}{4pt}
\caption{\define{Evaluation Results of Other Information Reducing Operators} We report the $\text{CRPS}_{\text{sum}}$ value at \gctsg task. }
\label{ablation:conditional_GC_TSG}
\resizebox{\textwidth}{!}{%
\begin{tabular}{l|ccc|ccc}
\toprule
& \multicolumn{3}{c|}{\textbf{ECG5000,  C2F $i\!\rightarrow\!j$}} 
& \multicolumn{3}{c}{\textbf{Nasdaq,  C2F $i\!\rightarrow\!j$}}\\
\cmidrule(lr){2-4}\cmidrule(lr){5-7}
\textbf{Operator}
& \textbf{TimeGAN}
& \textbf{TimeVQVAE}
& \textbf{\name}
& \textbf{TimeGAN}
& \textbf{TimeVQVAE}
& \textbf{\name} \\
\midrule
Wavelet
& 3.084\stdtiny{1.111}& 6.475\stdtiny{0.952}& \textbf{2.354\stdtiny{0.916}}& 39.80\stdtiny{22.00}& 35.68\stdtiny{34.24}& \textbf{17.29\stdtiny{13.05}}\\
Moving Average
& 2.702\stdtiny{1.199}& 3.884\stdtiny{0.852}& \textbf{2.080\stdtiny{0.905}}& 39.66\stdtiny{21.99}& 35.75\stdtiny{34.99}& \textbf{17.13\stdtiny{12.91}}\\
Downsample
& 2.686\stdtiny{1.126}& 3.496\stdtiny{0.998}& \textbf{2.126\stdtiny{0.915}}& 39.51\stdtiny{22.15}& 34.63\stdtiny{34.14}& \textbf{16.50\stdtiny{12.55}}\\
\midrule
Gaussian
& 1.737\stdtiny{0.737}& 3.693\stdtiny{0.858}& \textbf{1.587\stdtiny{0.867}}& 39.11\stdtiny{22.14}& 35.32\stdtiny{34.30}& \textbf{15.87\stdtiny{12.48}}\\
\bottomrule
\end{tabular}
}%
\end{table*}

In~\cref{ablation:conditional_GC_TSG}, we evaluate the \gctsg performance of \name on evaluation sets constructed with various information reducing operation functions~($\mathcal{R}_{\ell}$). Although \name is trained exclusively with Gaussian kernel smoothing, it generalizes robustly to other operators such as wavelet transforms, moving averages, and down sampling. This robustness is particularly important for practical applications, as it demonstrates that \name can effectively handle diverse coarse inputs conditions, such as user provided sketches and coarse outline.

\newpage
\section{Use Case Analysis}
\label{appendix:use_case}
\begin{figure*}[!ht]
\begin{center}
\center{\includegraphics[width=\textwidth]{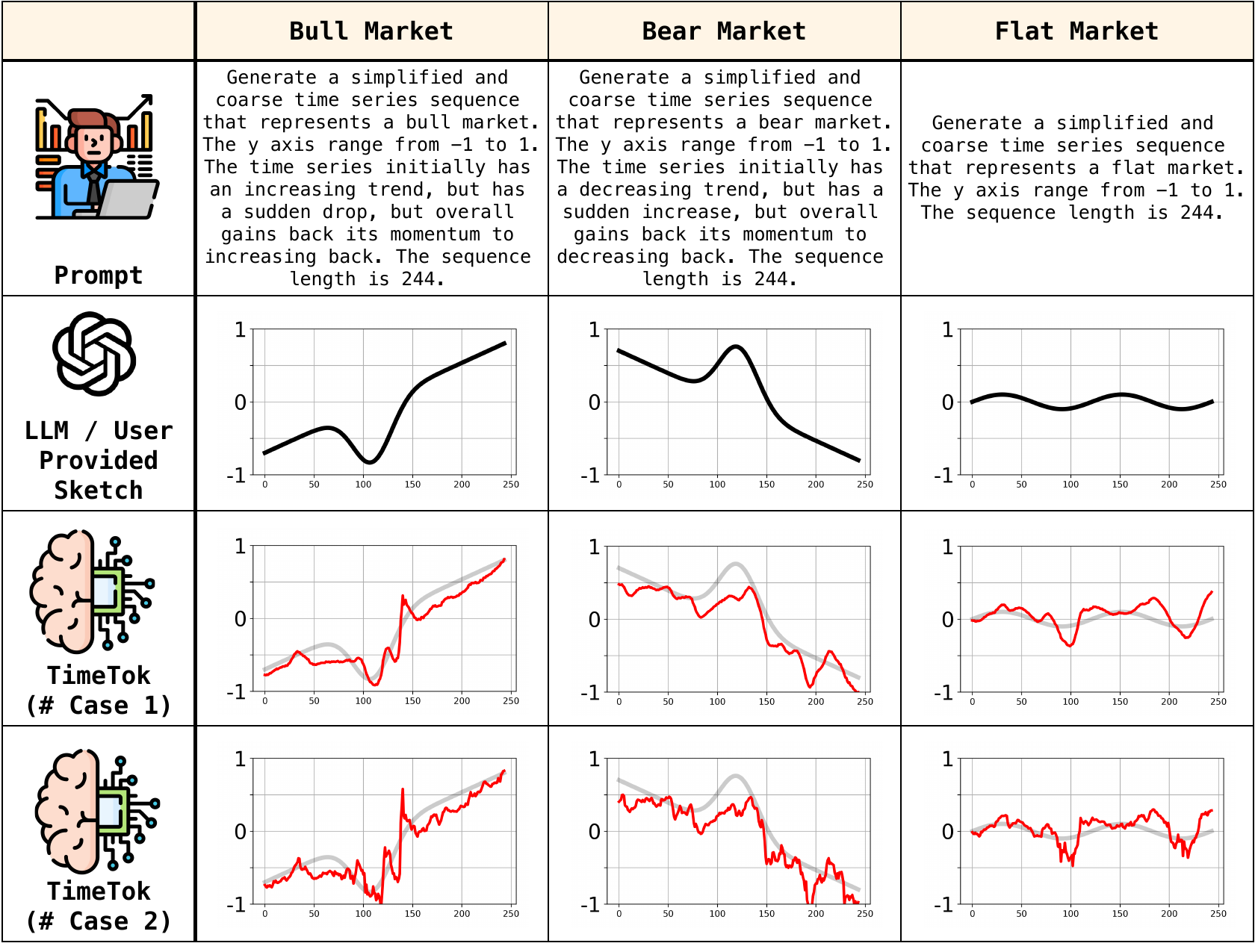}}
\caption{\define{Generating Fine-Grained Signals Conditioned on Coarse Outline: An Analysis in Financial Market Scenario} For this analysis, we used \name trained on the Nasdaq dataset. We employed ChatGPT to write code that generates the coarse signal, then used \name to generate time series at various granularity levels, conditioned on the given coarse signal. }
\label{appendix_fig:usecase_analysis}
\end{center}
\end{figure*}

Complex time series are inherently difficult to describe using natural language due to their temporal structure and inherent randomness. In contrast, specifying a coarse outline of the time series is relatively straightforward: a user can simply draw the outline or ask an LLM to generate such signals or scenarios. This motivates an important application of \name: generating fine-grained signals conditioned on a coarse outline. For instance, a financial analyst may want to model a bull market scenario, but with a slight drop in between. Such a coarse outline can be easily provided with an LLM as shown in the bull market scenario in~\cref{appendix_fig:usecase_analysis}. Using this coarse outline, the analyst can use \name to generate realistic financial bear market movement at various granularity levels, conditioned on the signal. We note that such generation is grounded on the actual distribution of the stock market~(\ie \name's training data), differentiating with methods that simply add random noise to the coarse structure.   

\section{The Use of LLMs in this work}
\label{app:use_of_llms}

In this paper, we employed Large Language Models (LLMs) for sentence restructuring to enhance clarity. We also used LLMs for the experiments in~\cref{appendix:use_case}.


\newpage
\section{Token Usage Analysis on the Nasdaq Dataset}
\label{appendix:token_usage}
\begin{figure*}[!ht]
\begin{center}
\center{\includegraphics[width=\textwidth]{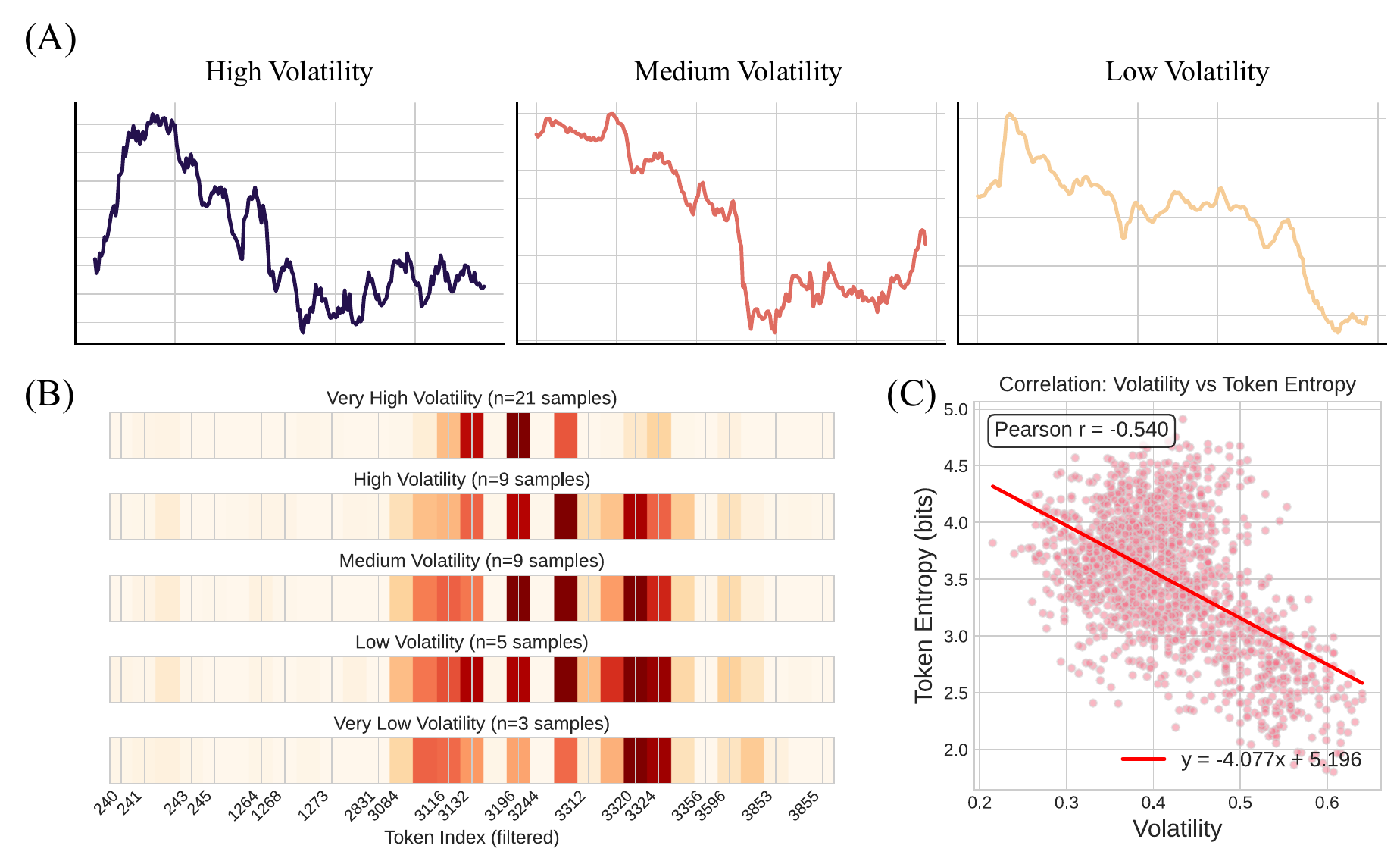}}
\caption{\define{Token Usage Pattern Analysis in Nasdaq} \textbf{(A)} We quantify samples based on their volatility, and visualize three representative samples. \textbf{(B)} To narrow down the analysis space, we filter to samples that start with token index \texttt{3260}, which generally exhibit an increasing trend pattern. From these samples, we visualize a heatmap depicting the token usage pattern at the full sequence length (\ie 128 tokens). We observe that higher-volatility samples tend to utilize fewer tokens, whereas lower-volatility samples exhibit larger and more diverse token usage. \textbf{(C)} Understanding that such pattern occurs in the selected samples, we subsequently expand our focus to all samples. In the scatter plot, the x-axis represents the volatility, and the y-axis is the token usage entropy. We observe that such token-usage pattern occurs throughout the whole samples.}
\label{appendix_fig:token_usage_analysis}
\end{center}
\end{figure*}

Our analysis focus on how the tokens in \name are used to represent the given time series. To analyze token usage pattern in \name, we used the Nasdaq synthetic dataset generated for the marginal experiment~(the dataset used in \cref{tab:marginal_fid_nasdaq}). Specifically, we used samples generated at level 8~($\ell=8$), as they utilize the full tokens (\ie $M=128$). We describe our analysis procedure in three stages.

\define{(A)} Given the time series, we first quantify the volatility of each data samples by measuring the Turning Point Rate~\citep{box2015time}, measuring the jaggedness of the data. In~\cref{appendix_fig:token_usage_analysis}-(A), we visualize samples based on the quantified measure. High volatility samples have high jaggedness, while low volatility samples have low jaggedness.

\define{(B)} We then tokenize the time series. To focus our token analysis, we identify token index \texttt{3260} as the most frequently occurring first token ($\ell=1$) in the Nasdaq dataset and retain only the samples that begin with this token. Since the first token captures the coarsest pattern of the time series, we observe that token index \texttt{3260} generally corresponds to an increasing trend. From these samples, we visualize the used token indexes from all token sequences (\ie 128) using a heatmap. We observe that higher-volatility samples tend to use smaller subset of tokens, while lower-volatility samples have more diverse use of tokens.

\define{(C)} With this finding, we subsequently expand our analysis to the full sample space. We quantify the token usage pattern with token entropy, where a higher entropy value means more diverse use of tokens, and lower entropy corresponds to more selective use of tokens. From the scatter plot, we observe that a similar token usage pattern occurs throughout all samples.

In summary, we observe that samples with high volatility use a smaller subset of tokens, while lower volatility samples use a larger subset of tokens.

\newpage
\section{Robustness Between Evaluators}
\label{appendix:Robustness}

We evaluate TSTR performance across linear (DLinear), recurrent (xLSTM), convolutional (InceptionTime), Transformer, and pretrained foundation-model evaluators on ECG5000 and ETTh1. For foundation-model evaluation, we use TimEE for classification without gradient updates and Chronos-2 for forecasting with LoRA fine-tuning.
\name achieves mean generator ranks of 1.4 on ECG5000 and 3.2
on ETTh1, supporting the robustness of the main findings
across different evaluators.

\begin{table}[!htbp]
\centering
\caption{ECG5000 --- F1 (higher is better).}
\small
\setlength{\tabcolsep}{3pt}
\begin{tabular}{c|ccccc}
\hline
& \makecell{Linear\\(DLinear)}
& \makecell{RNN\\(xLSTM)}
& \makecell{CNN\\(InceptionTime)}
& Transformer
& \makecell{TS-FM\\(TIMEE)} \\
\hline
RNN (TF)     & 0.326$\pm$0.003 & 0.356$\pm$0.010 & 0.382$\pm$0.009 & 0.340$\pm$0.014 & 0.378$\pm$0.000 \\
TimeVQVAE    & 0.443$\pm$0.003 & 0.383$\pm$0.013 & 0.379$\pm$0.000 & 0.421$\pm$0.024 & 0.384$\pm$0.000 \\
DiffusionTS  & 0.567$\pm$0.031 & 0.481$\pm$0.039 & 0.533$\pm$0.054 & 0.422$\pm$0.042 & 0.562$\pm$0.000 \\
ImagenTime   & 0.626$\pm$0.054 & 0.504$\pm$0.037 & 0.488$\pm$0.093 & 0.554$\pm$0.034 & 0.608$\pm$0.000 \\
TimeTok      & 0.555$\pm$0.020 & 0.512$\pm$0.045 & 0.583$\pm$0.016 & 0.585$\pm$0.033 & 0.622$\pm$0.000 \\
Real         & 0.620$\pm$0.006 & 0.613$\pm$0.025 & 0.630$\pm$0.015 & 0.563$\pm$0.065 & 0.605$\pm$0.000 \\
\hline
\end{tabular}
\end{table}


\begin{table}[!htbp]
\centering
\caption{ECG5000 --- F1: method rank per evaluator (1 = best).}
\label{tab:ecg5000_f1_rank}
\small
\setlength{\tabcolsep}{3.5pt}
\begin{tabular}{c|ccccc|c}
\hline
& \makecell{Linear\\(DLinear)}
& \makecell{RNN\\(xLSTM)}
& \makecell{CNN\\(InceptionTime)}
& Transformer
& \makecell{TS-FM\\(TIMEE)}
& \makecell{Mean\\rank} \\
\hline
TimeTok     & 3 & 1 & 1 & 1 & 1 & 1.4 \\
ImagenTime  & 1 & 2 & 3 & 2 & 2 & 2.0 \\
DiffusionTS & 2 & 3 & 2 & 3 & 3 & 2.6 \\
TimeVQVAE   & 4 & 4 & 5 & 4 & 4 & 4.2 \\
RNN (TF)    & 5 & 5 & 4 & 5 & 5 & 4.8 \\
\hline
\end{tabular}
\end{table}

\begin{table}[!htbp]
\centering
\caption{ETTh1 --- MSE (lower is better).}
\small
\setlength{\tabcolsep}{3pt}
\begin{tabular}{c|ccccc}
\hline
& \makecell{Linear\\(DLinear)}
& \makecell{RNN\\(xLSTM)}
& \makecell{CNN\\(InceptionTime)}
& Transformer
& \makecell{TS-FM\\(Chronos-2)} \\
\hline
RNN (TF)     & 0.104$\pm$0.000 & 0.112$\pm$0.001 & 0.111$\pm$0.001 & 0.107$\pm$0.001 & 0.104$\pm$0.000 \\
TimeVQVAE    & 0.102$\pm$0.001 & 0.113$\pm$0.002 & 0.117$\pm$0.003 & 0.110$\pm$0.001 & 0.106$\pm$0.001 \\
DiffusionTS  & 0.101$\pm$0.001 & 0.105$\pm$0.001 & 0.109$\pm$0.004 & 0.102$\pm$0.001 & 0.102$\pm$0.000 \\
ImagenTime   & 0.099$\pm$0.000 & 0.073$\pm$0.002 & 0.013$\pm$0.003 & 0.090$\pm$0.005 & 0.104$\pm$0.000 \\
TimeTok      & 0.103$\pm$0.000 & 0.107$\pm$0.001 & 0.102$\pm$0.002 & 0.105$\pm$0.002 & 0.105$\pm$0.001 \\
Real         & 0.098$\pm$0.000 & 0.056$\pm$0.003 & 0.005$\pm$0.001 & 0.098$\pm$0.001 & 0.103$\pm$0.000 \\
\hline
\end{tabular}
\end{table}



\begin{table}[!htbp]
\centering
\caption{ETTh1 --- MSE: method rank per evaluator (1 = best).}
\label{tab:etth1_mse_rank}
\small
\setlength{\tabcolsep}{3.5pt}
\begin{tabular}{c|ccccc|c}
\hline
& \makecell{Linear\\(DLinear)}
& \makecell{RNN\\(xLSTM)}
& \makecell{CNN\\(InceptionTime)}
& Transformer
& \makecell{TS-FM\\(Chronos-2)}
& \makecell{Mean\\rank} \\
\hline
ImagenTime  & 1 & 1 & 1 & 1 & 2 & 1.2 \\
DiffusionTS & 2 & 2 & 3 & 2 & 1 & 2.0 \\
TimeTok     & 4 & 3 & 2 & 3 & 4 & 3.2 \\
RNN (TF)    & 5 & 4 & 4 & 4 & 3 & 4.0 \\
TimeVQVAE   & 3 & 5 & 5 & 5 & 5 & 4.6 \\
\hline
\end{tabular}
\end{table}





\end{document}